\documentclass{article}
\usepackage{caption}
\usepackage{microtype}
\usepackage{graphicx}
\usepackage{subcaption}
\usepackage{booktabs}
\usepackage{enumitem}
\usepackage{multirow}
\usepackage[most]{tcolorbox} \usepackage{tabularx}
\usepackage{hyperref}
\usepackage{fvextra}
\usepackage{tikz}
\usepackage{adjustbox}
\usepackage{ragged2e}
\usetikzlibrary{arrows.meta,positioning,shadows.blur,fit}

\usepackage[ruled,vlined,linesnumbered]{algorithm2e}
\usepackage{xcolor}
\usepackage[most]{tcolorbox}
\newtcolorbox{promptbox}[1]{
  enhanced,
  breakable,
  colback=gray!4,
  colframe=gray!55,
  title=#1,
  fonttitle=\bfseries,
  boxrule=0.5pt,
  arc=2pt
}
\definecolor{PromptFrame}{HTML}{4B5563}
\definecolor{PromptBack}{HTML}{F9FAFB}

\definecolor{CoTColor}{HTML}{2563EB}
\definecolor{CoTSCColor}{HTML}{7C3AED}
\definecolor{ToTColor}{HTML}{EA580C}
\definecolor{SelfRefineColor}{HTML}{059669}

\newtcolorbox{methodpromptbox}[3][]{
    enhanced,
    width=\columnwidth,
    colback=#2!4,
    colframe=#2!70,
    boxrule=0.45pt,
    arc=1.5pt,
    left=4pt,
    right=4pt,
    top=4pt,
    bottom=4pt,
    title={#3},
    fonttitle=\scriptsize\bfseries,
    coltitle=black,
    boxed title style={
        colback=#2!12,
        colframe=#2!70,
        boxrule=0.3pt,
        arc=1.5pt
    },
    attach boxed title to top left={xshift=4pt,yshift=-1.5pt},
    #1
}

\usepackage[accepted]{icml2026}

\usepackage{amsmath}
\usepackage{amssymb}
\usepackage{mathtools}
\usepackage{amsthm}

\usepackage[capitalize,noabbrev]{cleveref}

\theoremstyle{plain}

\theoremstyle{definition}

\theoremstyle{remark}

\usepackage[textsize=tiny]{todonotes}

\icmltitlerunning{ArtMine: Discovering and Formalizing Artistic Processes}

\begin{document}

\twocolumn[
\icmltitle{ArtMine: Discovering and Formalizing Artistic Processes}

\icmlsetsymbol{equal}{*}

\begin{icmlauthorlist}
\icmlauthor{Kaustubh Kumar}{tcs,iit}
\icmlauthor{Ashutosh Ranjan}{tcs}
\icmlauthor{Vivek Srivastava}{tcs}
\icmlauthor{Blessin Varkey}{tcs}
\icmlauthor{Shirish Karande}{tcs}

\end{icmlauthorlist}

\icmlaffiliation{tcs}{TCS Research, Pune, India}
\icmlaffiliation{iit}{Indian Institue of Technology, Patna, India}
\icmlcorrespondingauthor{Kaustubh Kumar}{2201mm14\_kaustubh@iitp.ac.in}
\icmlcorrespondingauthor{Ashutosh Ranjan}{ashutosh.ranjan2@tcs.com}
\icmlcorrespondingauthor{Vivek Srivastava}{srivastava.vivek2@tcs.com}
\icmlcorrespondingauthor{Blessin Varkey}{blessin.varkey@tcs.com}
\icmlcorrespondingauthor{Shirish Karande}{shirish.karande@tcs.com}

\icmlkeywords{Multi-Agent Systems, Abductive Reasoning, Human-AI Co-Creativity,
              Process Learning, Generative AI, Computational Art History}

\vskip 0.3in
]

\printAffiliationsAndNotice{}  

\begin{abstract}
Understanding how artworks are created requires reasoning about the iterative decisions, material operations, and contextual influences that shape artistic production. While recent generative AI systems can synthesize artworks with high fidelity, they primarily model distributions over finished artifacts rather than the creative processes underlying their creation. In practice, artistic workflows are only partially documented through fragmented sources such as archival records, preparatory studies, correspondence, etc., making process-level understanding difficult to formalize computationally. In this work, we introduce \textit{\textbf{ArtMine}}, a framework for discovering and formalizing artistic processes from heterogeneous historical evidence. Our approach synthesizes heterogeneous artwork evidence into a structured repository, from which a Peircean abductive agent infers evidence-grounded production steps. These steps are converted into a compositional graph and rendering prompt, then optimized through self-reflection over deviations between the generated and reference artworks. We provide a preliminary proof-of-concept case study using open-domain historical sources across multiple artists and artistic movements, demonstrating that fragmented documentary evidence can support coherent, interpretable, and auditable representations of artistic workflows. By modeling creative processes rather than only final artifacts, our work moves toward process-centered human-AI co-creativity systems that can support artistic interpretation, creative education, reflective collaboration, and computational studies of cultural production.

\end{abstract}

\section{Introduction}
\label{sec:intro}

Recent progress in generative AI has been primarily driven by learning distributions over completed artifacts such as images, text, music, and video \cite{brown2020language,ho2020denoising,rombach2022high}. These systems can reproduce the stylistic and semantic properties of human-created works with remarkable fidelity, but they largely model \textit{what} creative works look like rather than \textit{how} they are made. The sequence of decisions, revisions, material operations, and intermediate reasoning that shape creative practice typically remains latent.

This limitation matters for human-AI co-creativity, where collaboration often depends on intermediate processes rather than final outputs alone. Artists iteratively sketch, refine, critique, and revise their work, while current generative systems operate primarily through prompt-to-artifact generation. At the same time, substantial evidence about creative processes already exists outside of the artifact itself. Artist correspondence, preparatory sketches, conservation reports, provenance records, technical imaging, and scholarly critique collectively form a fragmented but information-rich record of artistic production. Human scholars routinely use such evidence to reconstruct plausible production trajectories and understand stylistic evolution. However, despite advances in research agents \cite{nakano2021webgpt,yao2022react,shinn2023reflexion} and computational creativity \cite{cohen1995further,colton2012painting}, these documentary records have rarely been treated as supervision to model creative processes.

In this work, we study \textit{artistic process discovery and formalization}: the task of inferring structured, auditable sequences of production steps for artworks from heterogeneous historical evidence. The objective is to recover plausible creative trajectories that explain how an artwork may have been produced, while preserving explicit links between inferred actions and their supporting sources. We frame this task as an abductive inference under incomplete information \cite{josephson1996abductive,peirce1997pragmatism,kakas1992abductive}.
Given an artwork and a heterogeneous corpus of documentary evidence, the system incrementally constructs hypotheses about the latent process that could plausibly have produced the work while remaining consistent with the available historical and material record.
\begin{figure*}[t]
\centering
\includegraphics[width=1.0\textwidth]{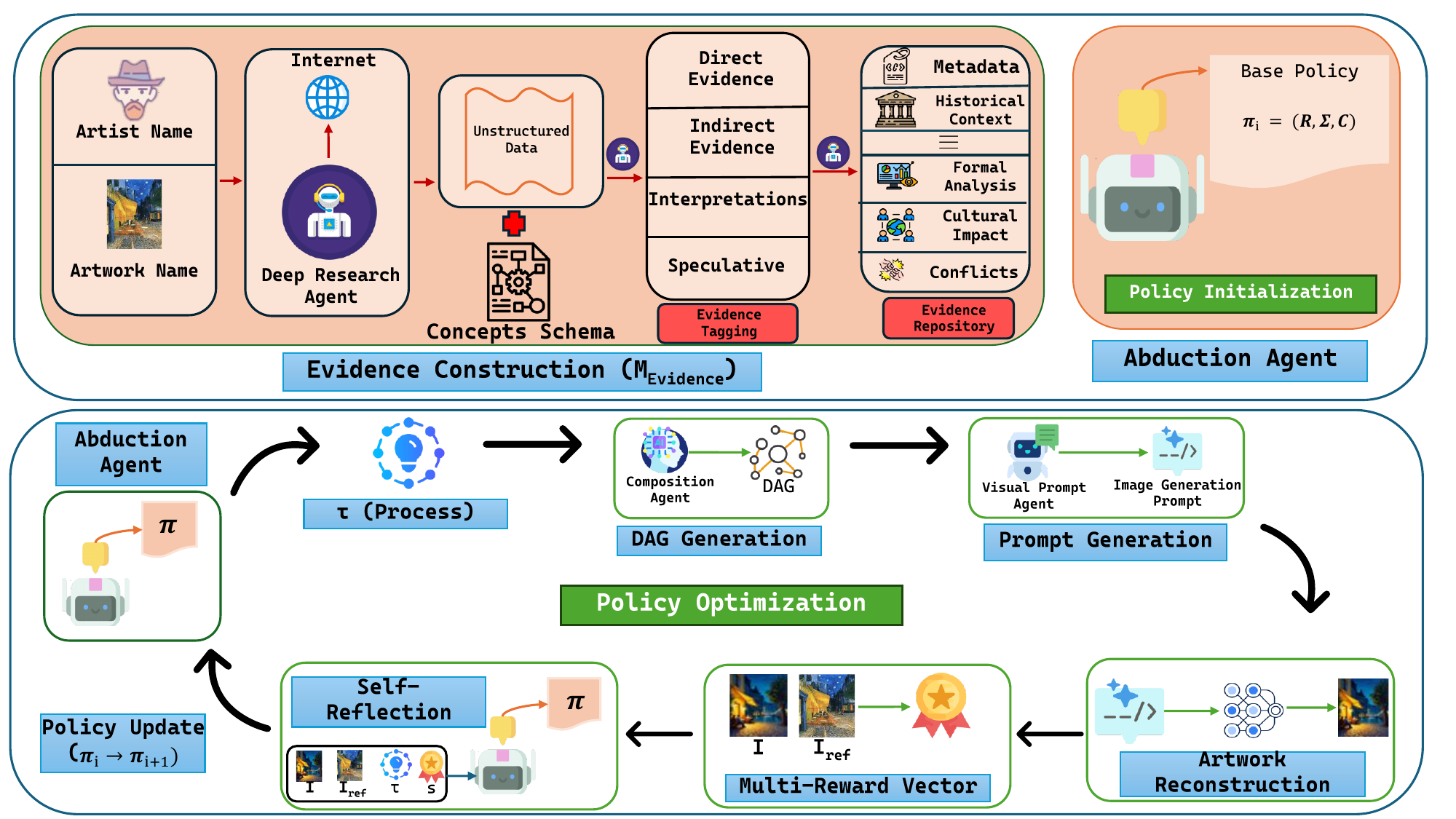}
\caption{
\textbf{\textit{ArtMine}}. 
Architecture diagram of our proposed framework for discovering and formalizing artistic processes.
}

\label{fig:architecture}
\end{figure*}
To address this task, we propose \textit{ArtMine}, a framework that integrates deep-research-based evidence construction, structured evidence repository, Peircean abductive reasoning, and self-reflection-driven policy updates for artistic process discovery.
 A deep-research agent organizes open-domain evidence into a structured concept schema. An abductive agent then uses this repository to infer process trajectories under a Peircean policy linking observations, rules, and hypothesized production actions. The inferred trajectory is rendered, compared with the reference artwork using multi-reward feedback, and iteratively improved through self-reflection-driven policy updates (see Figure \ref{fig:architecture}).
Our case study demonstrates that heterogeneous historical evidence can be organized into structured representations suitable for downstream reasoning, and that abductive reconstruction can generate coherent and auditable production trajectories grounded in those records. Our contributions are threefold:

\begin{enumerate}[noitemsep,nolistsep,leftmargin=*]
    \item We introduce \textbf{artistic process discovery and formalization} as a computational problem for generative AI, shifting the focus from modeling completed artifacts to reconstructing creative processes from heterogeneous historical evidence.
    
    \item We propose \textbf{\textit{ArtMine}} an agentic framework that combines deep-research-based evidence construction, Peircean abductive reasoning, and se to reconstruct auditable creative trajectories from unstructured documentary and archival records.
    
    \item We present a \textbf{case study} as proof-of-concept reconstructions demonstrating that open-domain documentary and evidence can support structured inference over creative processes.
\end{enumerate}

\section{Related Work}
\label{sec:related}


Recent research-agent systems extend retrieval-augmented generation with iterative planning, tool use, reflection, and web interaction \citep{lewis2020retrieval,yao2022react,shinn2023reflexion,nakano2021webgpt}. These systems demonstrate that language agents can gather and synthesize information from heterogeneous online sources through multi-step reasoning and interaction. Our work builds on this direction by using a Deep Research agent to gather evidence from multiple sources, synthesize claims across these sources, assign reliability tags to retrieved evidence, and organize the resulting information into a structured evidence repository for downstream abductive reasoning.
Rather than treating retrieved information as unstructured context for generation, the resulting evidence is organized into structured representations that support downstream reasoning about creative processes.

Role-conditioned and persona-conditioned language agents aim to maintain behavioral or stylistic consistency during generation. Early work represented personas through static textual profiles in dialogue systems \citep{zhang2018personalizing,mazare2018training}, while more recent approaches study role-playing and character-conditioned generation in large language models. ArtPeer \citep{kumarartpeer} explores artist-centered persona construction from art-historical sources, while CharacterGLM \citep{zhou2024characterglm} studies character-consistent generation under persona conditioning. In most of these systems, personas primarily function as conditioning mechanisms for generating stylistically or behaviorally consistent outputs. In contrast, our framework evaluates discovered process trajectories through artwork-grounded policy optimization, where rendered artwork are compared with reference artworks and the resulting deviations drive self-reflection-based policy updates.

Our work is also related to computational creativity and generative art systems. Classical systems such as AARON \citep{cohen1995further} and The Painting Fool \citep{colton2012painting} generate artworks through procedural rules and heuristic creative strategies. More recent generative models such as DALL$\cdot$E~3 \citep{betker2023improving}, Stable Diffusion \citep{rombach2022high}, and Jukebox \citep{dhariwal2020jukebox} learn distributions over artistic artifacts from large-scale multimodal data. Agent-based simulations of social behavior and interaction \citep{park2023generative} similarly model emergent behaviors through multi-agent interaction. In contrast, our work focuses on discovering historically grounded creative trajectories from documentary, archival, technical, and art-historical evidence, and refining them through artwork-grounded self-reflection rather than treating generation as a one-shot artifact-production task. A detailed discussion of the related works is presented in the Appendix \ref{sec:ext_related}.

\section{Method}
\label{sec:method}

\subsection{Evidence Construction}
\label{sec:evidence}

Discovering how an artwork was made requires more than describing its visible surface. The finished object shows the result, not the order of actions, revisions, abandoned alternatives, or decisions that produced it, and since different processes can yield similar effects, the object alone does not determine its own history. Process discovery therefore requires external knowledge: artist letters and preparatory studies \cite{van2009vincent}, imaging and pigment studies \citep{zhao2008investigation,delaney2016visible}, museum and auction records, and later scholarship. These sources constrain what is plausible, as technical studies reveal hidden layers and materials, correspondence records intent and working conditions, and preparatory studies expose earlier states. But they differ in reliability: a statement by the artist, a conservation report, and a speculative claim cannot be treated alike, and once retrieved as plain text, these differences are easily hidden. Since downstream reasoning depends on the reliability of evidence
\citep{shi2023large}, evidence typing must occur while knowledge is gathered, not after it.

We construct the evidence repository $\mathcal{M}_{\text{evidence}}$ using a Deep Research agent that plays two roles. First, it retrieves information about the artwork from open-domain sources, such as museum collection records, conservation and technical reports, artist correspondence, archival documents, exhibition catalogs, scholarly articles, provenance records, etc. Second, it organizes that information into a concept schema with 11 dimensions: \textbf{metadata, historical context, materials and technique, formal analysis, iconographic analysis, documentary evidence, provenance, scientific analysis, interpretations, cultural impact, and conflicts}. This schema is designed to capture both iconographic and technical aspects of an artwork. The iconographic dimensions focus on visual content, symbolic meaning, and interpretation \cite{panofsky2018studies}, while the technical dimensions record materials, methods, condition, provenance, and scientific findings \cite{hermens2012technical}. To support reasoning under uncertainty, the repository also preserves disagreement between sources. When sources offer conflicting claims, the agent does not collapse them into a single reconciled statement; instead, it stores the competing positions in a dedicated \texttt{conflicts} category. This allows the abductive solver to compare alternative explanations while maintaining an explicit awareness of uncertainty in the evidence.

\begin{figure}[t]
  \centering
  \includegraphics[width=0.95\linewidth]{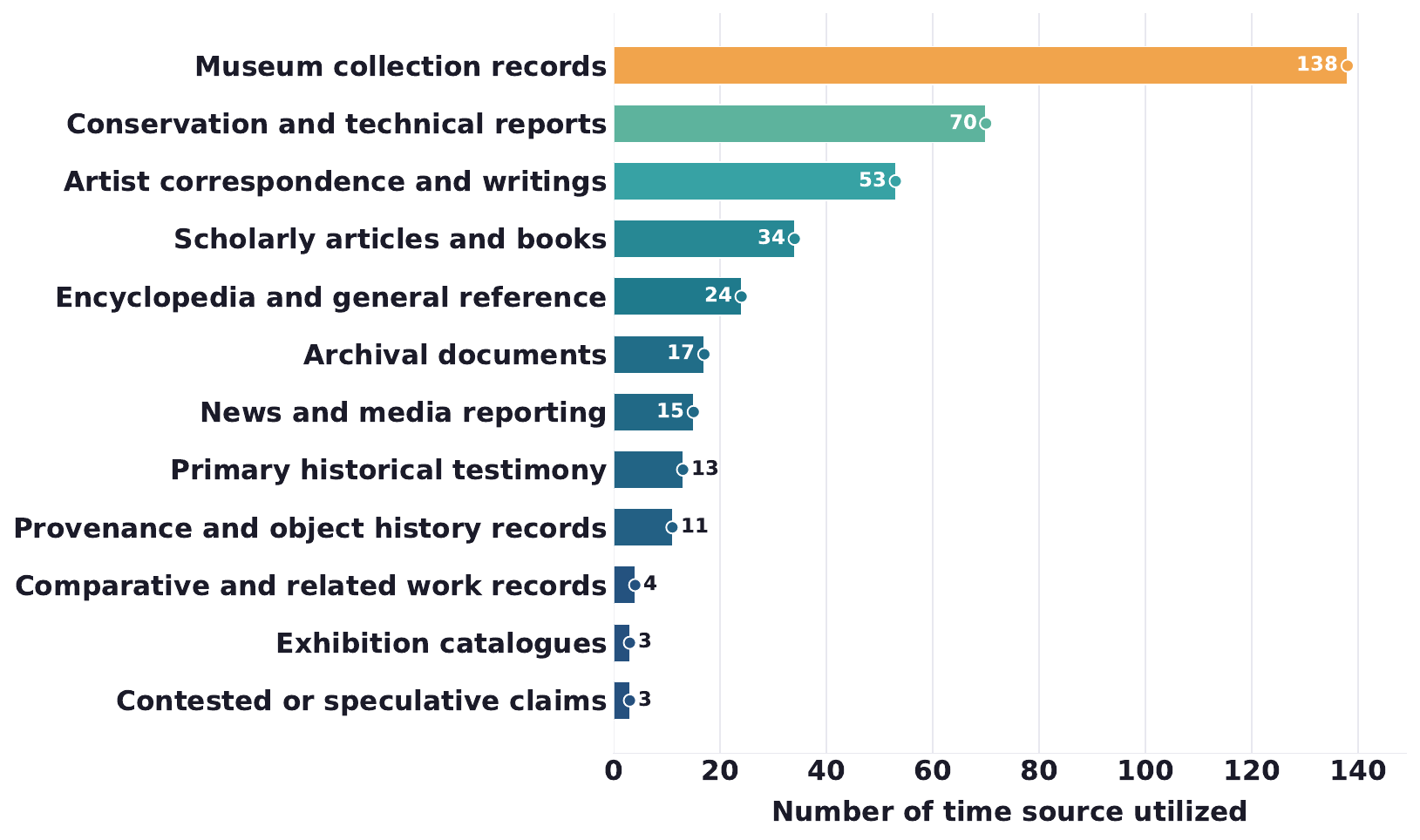}
  \caption{Distribution of evidence sources across $M_{evidence}$, counted
  as the number of times each source category is utilized across all evidence
  assertions in our case study.}
  \label{fig:source-categories}
\end{figure}


The evidence in the repository is tagged by how it was obtained: the Deep Research
agent assigns each piece of evidence a tag based on its source and method of
discovery, either \texttt{direct\_evidence}, \texttt{indirect\_evidence},
\texttt{interpretation}, or \texttt{speculative} \cite{howell2001reliable}.
Here, \texttt{direct\_evidence} refers to primary sources, such as artist letters, archival records, museum entries, conservation
reports, etc. \texttt{Indirect\_evidence} captures claims
inferred from contextual or secondary sources, including biographies,
exhibition histories, related artworks, or known workshop practices.
\texttt{Interpretation} denotes scholarly readings of meaning, symbolism,
style, or intention, while \texttt{speculative} marks low-certainty claims that
lack strong documentary or material support. These distinctions allow the
abduction agent to reason with graded reliability instead of treating all
retrieved claims as equally trustworthy. For example, a
signed letter by the artist and a speculative claim from popular writing then
carry different weight in reasoning, rather than being treated as equally
reliable. 

The agent is also instructed not to invent missing information or merge the interpretation with fact. As a result, $\mathcal{M}_{\text{evidence}}$ is not a flat text summary but a structured evidence base where each claim carries its source, reliability, and uncertainty. An example $\mathcal{M}_{\text{evidence}}$ for \textit{Mona Lisa} is provided in the Appendix \ref{app:memory-monalisa}.


\definecolor{CodeBack}{HTML}{FFFFFF}
\definecolor{CodeFrame}{HTML}{111111}
\newcommand{\PiZero}{\ensuremath{\Pi_{0}}}

\begin{figure}[t]
\centering
\vspace{-0.0em}

\begin{tcolorbox}[
    enhanced,
    width=\columnwidth,
    colback=CodeBack,
    colframe=CodeFrame,
    boxrule=0.55pt,
    arc=0pt,
    left=4pt,
    right=4pt,
    top=1pt,
    bottom=2pt,
    before skip=0pt,
    after skip=0pt
]

\begin{Verbatim}[
fontsize=\scriptsize,
commandchars=\\\{\},
breaklines=true,
breakanywhere=false,
breakautoindent=true,
breakindent=1.2em,
breaksymbolleft={},
breaksymbolright={},
baselinestretch=0.78,
tabsize=4
]
====================================================
Base Policy \PiZero
====================================================

def solver(agent, evidence_repository, artwork, 
policy=\PiZero):
    role = "art historian specializing in process reconstruction"

    schema = PEIRCEAN_SCHEMA(
        \textbf{OBSERVATION}: A directly evidenced 
        feature of the artwork or its context; 
        something that can be pointed to and verified.

        \textbf{RULE}: A general principle stating 
        that if process P were followed, observation 
        O would result.

        \textbf{HYPOTHESIS}: Therefore, process P was 
        most likely undertaken. State this as a 
        concrete, actionable instruction.

        \textbf{ACTION}: A concrete instruction 
        realizing the hypothesis; specifies what to 
        lay in or how to edit the current canvas.

        \textbf{EVIDENCE}: The evidence-repository 
        unit that grounds the observation, with 
        reliability tag: direct, indirect,
        interpretation, or speculative.
    )

    constraints = (
        "1. Apply Peircean schema for every step: "
        "observation, rule, hypothesis.\textbackslash n"
        "2. Every step must cite at least one specific key "
        "from the evidence JSON.\textbackslash n"
        "3. Steps must appear in production order; "
        "step 1 is first, step N is last.\textbackslash n"
        "4. Return valid JSON only.\textbackslash n"
    )

    return trajectory
\end{Verbatim}

\end{tcolorbox}

\caption{
\textbf{Base policy $\Pi_0$.}
The policy specifies the solver role, the Peircean schema used to structure abductive reasoning, and the requirements for generating grounded process-reconstruction steps. The solver is adapted from the frameworks of \cite{kakade2026polaris,yin2025godel}, with modifications to support evidence-grounded abductive process discovery in our setting.
}
\label{fig:base_policy}
\vspace{-0.12em}

\end{figure}
\definecolor{CodeBack}{HTML}{FFFFFF}
\definecolor{CodeFrame}{HTML}{111111}
\definecolor{UpdateGreen}{HTML}{059669}

\newcommand{\update}[1]{\textcolor{UpdateGreen}{#1}}
\newcommand{\PiThree}{\ensuremath{\Pi_{3}}}

\begin{figure}[t]
\centering

\begin{tcolorbox}[
    enhanced,
    width=\columnwidth,
    colback=CodeBack,
    colframe=CodeFrame,
    boxrule=0.55pt,
    arc=0pt,
    left=4pt,
    right=4pt,
    top=1pt,
    bottom=2pt,
    before skip=0pt,
    after skip=0pt
]

\begin{Verbatim}[
fontsize=\tiny,
commandchars=\\\{\},
breaklines=true,
breakanywhere=false,
breakautoindent=true,
breakindent=1.2em,
breaksymbolleft={},
breaksymbolright={},
baselinestretch=0.82,
tabsize=4
]
============================================================
Updated Policy \PiThree
============================================================

def solver(agent, evidence_repository, artwork, policy=\PiThree):
    role = "art historian specializing in process reconstruction"

    schema = PEIRCEAN_SCHEMA(
        \textbf{OBSERVATION}: A directly evidenced 
        feature of the artwork or its context; 
        something that can be pointed to and verified.

        \textbf{RULE}: A general principle stating 
        that if process P were followed, observation 
        O would result.

        \textbf{HYPOTHESIS}: The inferred production 
        explanation that identifies process P as the 
        most plausible cause of the observation.

        \textbf{ACTION}: A concrete instruction 
        realizing the hypothesis; specifies what to 
        lay in or how to edit the current canvas.

        \textbf{EVIDENCE}: The evidence-repository 
        unit that grounds the observation, with 
        reliability tag: direct, indirect, 
        interpretation, or speculative.
    )

    \update{method_principles = (}
        \update{"1. Render the central visual anchor at the scale"}
        \update{"   and position implied by the composition."}
        \update{"2. Define each element's palette using documented"}
        \update{"   color temperature and saturation."}
        \update{"3. Layer elements following the dark-to-light"}
        \update{"   sequence in the technique notes."}
        \update{"..."}
    \update{)}

    constraints = (
        "1. Apply Peircean schema for every step: "
        "observation, rule, hypothesis, action.\textbackslash n"
        "2. Every step must cite at least one specific key "
        "from the evidence JSON.\textbackslash n"
        "3. Steps must appear in production order; "
        "step 1 is first, step N is last.\textbackslash n"
        "4. Return valid JSON only.\textbackslash n"
        \update{"5. Refine the background with mountains, watercourses,"}
        \update{"   road, bridge, and sfumato transitions.\textbackslash n"}
        \update{"6. Adjust the smile and mouth to be softer and"}
        \update{"   more ambiguous.\textbackslash n"}
        \update{"7. Model the eyes with soft tonal gradations,"}
        \update{"   highlights, and shadow transitions.\textbackslash n"}
        \update{"8. Place the hands lower in the foreground and"}
        \update{"   render costume details faithfully.\textbackslash n"}
        \update{"...\textbackslash n"}
    )

    return trajectory
\end{Verbatim}

\end{tcolorbox}

\caption{
\textbf{Updated policy $\Pi_3$ after reflection.}
Black text denotes unchanged policy content, while green text marks learned method principles and newly added reconstruction requirements.
}
\label{fig:updated_policy}
\vspace{-0.8em}

\end{figure}

\subsection{Abductive Process Discovery}
\label{sec:abductive}

The evidence repository gives a structured account of what is known about an artwork but does not by itself explain how the artwork was made. It contains claims about materials, visual features, documents, provenance, and technical findings, but these claims must still be turned into a process. We therefore use an abductive agent to move from evidence to process. The goal is to infer a trajectory $\tau$ that explains how a blank canvas could have been transformed into an artwork consistent with the evidence in $\mathcal{M}_{\text{evidence}}$.

Abduction is appropriate for this setting because the process is not observed directly. The agent observes the finished artwork and the records around it, and tries to abduce which sequence of actions would best explain them \citep{josephson1996abductive,kapitan1992peirce,hoffman2020explaining}. The deduction is too strict for this task since no general rule entails the making of a particular painting. Induction is also limited, since we do not have a large set of comparable making processes from which to learn a reliable pattern. Abduction, instead, reasons backwards from observed evidence to a plausible explanation. Although recent work shows that language models can perform abductive inference \citep{bhagavatula2019abductive}, our setting requires the inference to remain grounded in an external evidence repository.

\definecolor{paperbg}{RGB}{250,247,240}
\definecolor{cream}{RGB}{255,250,235}
\definecolor{warmbrown}{RGB}{92,62,40}
\definecolor{softgold}{RGB}{239,200,130}
\definecolor{softrose}{RGB}{220,145,140}
\definecolor{mutedgreen}{RGB}{75,92,67}

\definecolor{dagpeach}{RGB}{247,201,158}
\definecolor{dagyellow}{RGB}{253,229,146}
\definecolor{dagpink}{RGB}{226,143,145}
\definecolor{dagline}{RGB}{150,125,20}
\definecolor{dagorange}{RGB}{255,160,0}
\definecolor{dagred}{RGB}{230,0,0}

\tcbset{
    panelbox/.style={
        enhanced,
        colback=paperbg,
        colframe=warmbrown!55,
        boxrule=0.65pt,
        arc=4pt,
        left=3pt,
        right=3pt,
        top=3pt,
        bottom=3pt,
        drop shadow={black!12!white},
        height=0.315\textheight,
        valign=center
    },
    innerbox/.style={
        colback=cream,
        colframe=warmbrown!40,
        boxrule=0.4pt,
        arc=3pt,
        left=3pt,
        right=3pt,
        top=3pt,
        bottom=3pt
    }
}

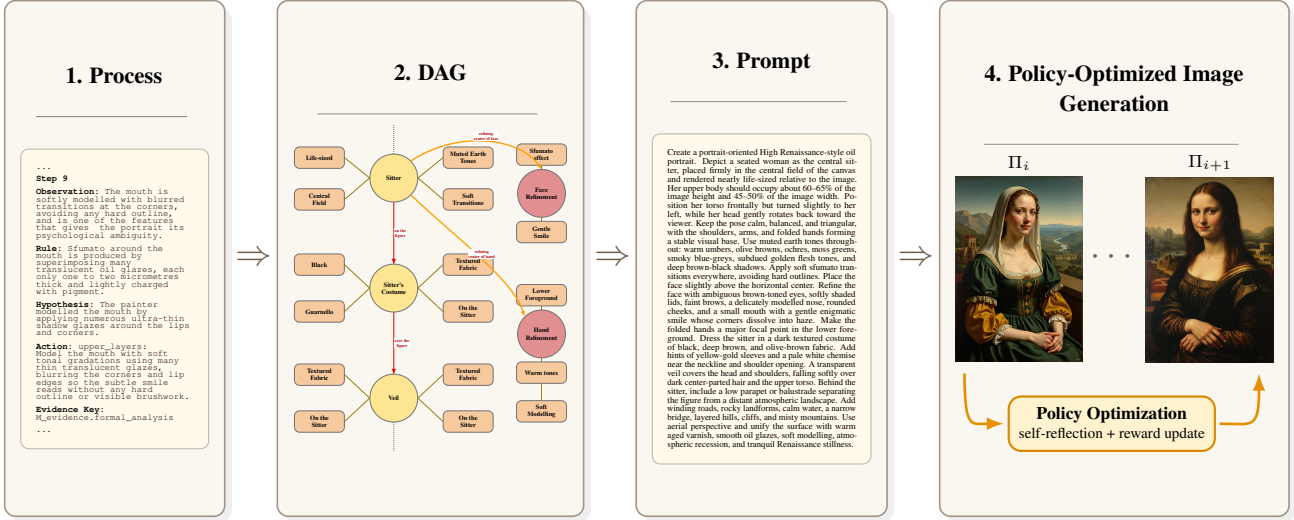
\begin{figure*}[t]
\centering

\setlength{\tabcolsep}{2pt}
\renewcommand{\arraystretch}{1.0}

\begin{adjustbox}{max width=\textwidth}
\begin{tabular}{
    >{\centering\arraybackslash}m{0.180\textwidth}
    >{\centering\arraybackslash}m{0.025\textwidth}
    >{\centering\arraybackslash}m{0.250\textwidth}
    >{\centering\arraybackslash}m{0.025\textwidth}
    >{\centering\arraybackslash}m{0.205\textwidth}
    >{\centering\arraybackslash}m{0.025\textwidth}
    >{\centering\arraybackslash}m{0.130\textwidth}
    >{\centering\arraybackslash}m{0.025\textwidth}
    >{\centering\arraybackslash}m{0.130\textwidth}
}

\begin{tcolorbox}[panelbox]
\centering
{\bfseries\small 1. Process}

\vspace{0.5mm}
{\color{warmbrown!60}\rule{0.80\linewidth}{0.45pt}}
\vspace{1mm}

\begin{tcolorbox}[innerbox]
\begin{Verbatim}[
fontsize=\fontsize{3.5}{3.55}\selectfont,
commandchars=\\\{\},
breaklines=true,
breakanywhere=false,
breakautoindent=true,
breakindent=1.1em,
breaksymbolleft={},
breaksymbolright={},
baselinestretch=0.74
]
...

\textbf{Step 9}

\textbf{Observation:} The mouth is 
softly modelled with blurred 
transitions at the corners, 
avoiding any hard outline, 
and is one of the features 
that gives  the portrait its 
psychological ambiguity.

\textbf{Rule:} Sfumato around the 
mouth is produced by 
superimposing many 
translucent oil glazes, each
only one to two micrometres
thick and lightly charged 
with pigment.

\textbf{Hypothesis:} The painter 
modelled the mouth by 
applying numerous ultra-thin 
shadow glazes around the lips
and corners.

\textbf{Action:} upper_layers: 
Model the mouth with soft
tonal gradations using many 
thin translucent glazes, 
blurring the corners and lip 
edges so the subtle smile 
reads without any hard 
outline or visible brushwork.

\textbf{Evidence Key:} 
M_evidence.formal_analysis

...
\end{Verbatim}
\end{tcolorbox}

\end{tcolorbox}

&
{\Large\color{warmbrown!85}$\Rightarrow$}
&

\begin{tcolorbox}[panelbox]
\centering
{\bfseries\small 2. DAG}

\vspace{0.5mm}
{\color{warmbrown!60}\rule{0.80\linewidth}{0.35pt}}
\vspace{1mm}

\begin{adjustbox}{width=0.99\linewidth,max height=0.285\textheight}
\begin{tikzpicture}[
    every node/.style={transform shape},
    box/.style={
        draw=black!55,
        fill=dagpeach,
        rounded corners=7pt,
        line width=0.45pt,
        align=center,
        font=\bfseries\small,
        minimum width=2.45cm,
        minimum height=1.05cm,
        inner sep=3pt
    },
    circlemain/.style={
        draw=black!55,
        fill=dagyellow,
        circle,
        line width=0.55pt,
        align=center,
        font=\bfseries\small,
        minimum size=2.35cm,
        inner sep=2pt
    },
    circlerefine/.style={
        draw=black!55,
        fill=dagpink,
        circle,
        line width=0.55pt,
        align=center,
        font=\bfseries\small,
        minimum size=2.35cm,
        inner sep=2pt
    },
    thinlink/.style={
        draw=dagline,
        line width=0.42pt
    },
    redarrow/.style={
        -{Latex[length=2.4mm,width=1.8mm]},
        draw=dagred,
        line width=0.75pt
    },
    orangearrow/.style={
        -{Latex[length=2.8mm,width=2.2mm]},
        draw=dagorange,
        line width=1.9pt
    },
    edgelabel/.style={
        fill=white,
        text=dagred!80!black,
        font=\bfseries\tiny,
        inner sep=0.8pt,
        align=center
    }
]

\node[circlemain] (sitter)  at (0, 5.4) {Sitter};
\node[circlemain] (costume) at (0, 0.0) {Sitter's\\Costume};
\node[circlemain] (veil)    at (0,-5.4) {Veil};

\node[box] (lifesized) at (-3.65, 6.40) {Life-sized};
\node[box] (central)   at (-3.65, 4.35) {Central\\Field};
\node[box] (earth)     at ( 3.65, 6.40) {Muted Earth\\Tones};
\node[box] (soft)      at ( 3.65, 4.35) {Soft\\Transitions};

\node[box] (black)     at (-3.65, 1.15) {Black};
\node[box] (guarnello) at (-3.65,-1.15) {Guarnello};
\node[box] (fabric1)   at ( 3.65, 1.15) {Textured\\Fabric};
\node[box] (onsitter1) at ( 3.65,-1.15) {On the\\Sitter};

\node[box] (fabric2)   at (-3.65,-4.25) {Textured\\Fabric};
\node[box] (onsitter2) at (-3.65,-6.55) {On the\\Sitter};
\node[box] (fabric3)   at ( 3.65,-4.25) {Textured\\Fabric};
\node[box] (onsitter3) at ( 3.65,-6.55) {On the\\Sitter};

\node[circlerefine] (face) at (7.25, 4.65) {Face\\Refinement};
\node[box] (sfumato)      at (7.25, 6.55) {Sfumato\\effect};
\node[box] (smile)        at (7.25, 2.75) {Gentle\\Smile};

\node[circlerefine] (hand) at (7.25,-2.25) {Hand\\Refinement};
\node[box] (lower)        at (7.25,-0.35) {Lower\\Foreground};
\node[box] (warm)         at (7.25,-4.15) {Warm tones};
\node[box] (model)        at (7.25,-6.05) {Soft\\Modelling};

\draw[thinlink] (lifesized.east) -- (sitter.west);
\draw[thinlink] (central.east) -- (sitter.west);
\draw[thinlink] (earth.west) -- (sitter.east);
\draw[thinlink] (soft.west) -- (sitter.east);

\draw[thinlink] (black.east) -- (costume.west);
\draw[thinlink] (guarnello.east) -- (costume.west);
\draw[thinlink] (fabric1.west) -- (costume.east);
\draw[thinlink] (onsitter1.west) -- (costume.east);

\draw[thinlink] (fabric2.east) -- (veil.west);
\draw[thinlink] (onsitter2.east) -- (veil.west);
\draw[thinlink] (fabric3.west) -- (veil.east);
\draw[thinlink] (onsitter3.west) -- (veil.east);

\draw[thinlink] (sfumato.south) -- (face.north);
\draw[thinlink] (face.south) -- (smile.north);

\draw[thinlink] (lower.south) -- (hand.north);
\draw[thinlink] (hand.south) -- (warm.north);
\draw[thinlink] (warm.south) -- (model.north);

\draw[redarrow]
    (sitter.south) -- 
    node[edgelabel, right, midway] {on the\\figure}
    (costume.north);

\draw[redarrow]
    (costume.south) -- 
    node[edgelabel, right, midway] {over the\\figure}
    (veil.north);

\draw[orangearrow]
    (sitter.north east)
    .. controls (3.2,7.8) and (6.2,7.35) ..
    node[edgelabel, above, midway] {refining\\center of face}
    (face.north);

\draw[orangearrow]
    (sitter.south east)
    .. controls (3.2,2.35) and (5.9,0.35) ..
    node[edgelabel, above, midway] {refining\\center of hand}
    (hand.north west);

\draw[dash pattern=on 1.2pt off 1.6pt, line width=0.8pt, draw=black!65]
    (sitter.north) -- ++(0,1.45);

\draw[dash pattern=on 1.2pt off 1.6pt, line width=0.8pt, draw=black!65]
    (veil.south) -- ++(0,-1.45);

\end{tikzpicture}
\end{adjustbox}

\end{tcolorbox}

&
{\Large\color{warmbrown!85}$\Rightarrow$}
&

\begin{tcolorbox}[panelbox]
\centering
{\bfseries\small 3. Prompt}

\vspace{0.5mm}
{\color{warmbrown!60}\rule{0.80\linewidth}{0.35pt}}
\vspace{1mm}

\begin{tcolorbox}[innerbox]
\fontsize{3.5}{3.5}\selectfont
Create a portrait-oriented High Renaissance-style oil portrait. Depict a seated woman as the central sitter, placed firmly in the central field of the canvas and rendered nearly life-sized relative to the image. Her upper body should occupy about 60--65\% of the image height and 45--50\% of the image width. Position her torso frontally but turned slightly to her left, while her head gently rotates back toward the viewer. Keep the pose calm, balanced, and triangular, with the shoulders, arms, and folded hands forming a stable visual base. Use muted earth tones throughout: warm umbers, olive browns, ochres, moss greens, smoky blue-greys, subdued golden flesh tones, and deep brown-black shadows. Apply soft sfumato transitions everywhere, avoiding hard outlines. Place the face slightly above the horizontal center. Refine the face with ambiguous brown-toned eyes, softly shaded lids, faint brows, a delicately modelled nose, rounded cheeks, and a small mouth with a gentle enigmatic smile whose corners dissolve into haze. Make the folded hands a major focal point in the lower foreground. Dress the sitter in a dark textured costume of black, deep brown, and olive-brown fabric. Add hints of yellow-gold sleeves and a pale white chemise near the neckline and shoulder opening. A transparent veil covers the head and shoulders, falling softly over dark center-parted hair and the upper torso. Behind the sitter, include a low parapet or balustrade separating the figure from a distant atmospheric landscape. Add winding roads, rocky landforms, calm water, a narrow bridge, layered hills, cliffs, and misty mountains. Use aerial perspective and unify the surface with warm aged varnish, smooth oil glazes, soft modelling, atmospheric recession, and tranquil Renaissance stillness.
\end{tcolorbox}

\end{tcolorbox}

&
{\Large\color{warmbrown!85}$\Rightarrow$}
&

\multicolumn{3}{>{\centering\arraybackslash}m{0.285\textwidth}}{

\begin{tcolorbox}[panelbox]
\centering
{\bfseries\small 4. Policy-Optimized Image Generation}

\vspace{0.5mm}
{\color{warmbrown!60}\rule{0.80\linewidth}{0.35pt}}
\vspace{1mm}

\begin{tabular}{@{}c@{\hspace{1mm}}c@{\hspace{1mm}}c@{}}

\begin{minipage}[c]{0.40\linewidth}
\centering
{\scriptsize $\Pi_i$}

\vspace{0.5mm}

\includegraphics[
    width=\linewidth,
    height=0.215\textheight,
    keepaspectratio
]{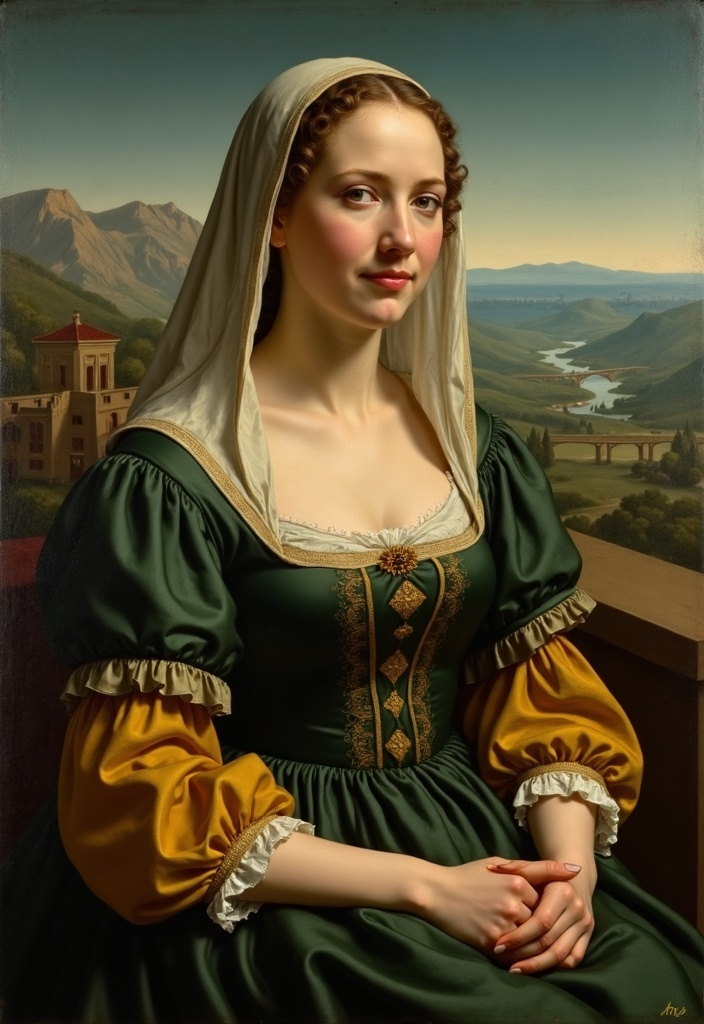}
\end{minipage}

&

{\Large\bfseries\color{warmbrown!80}$\cdots$}

&

\begin{minipage}[c]{0.40\linewidth}
\centering
{\scriptsize $\Pi_{i+1}$}

\vspace{0.5mm}

\includegraphics[
    width=\linewidth,
    height=0.215\textheight,
    keepaspectratio
]{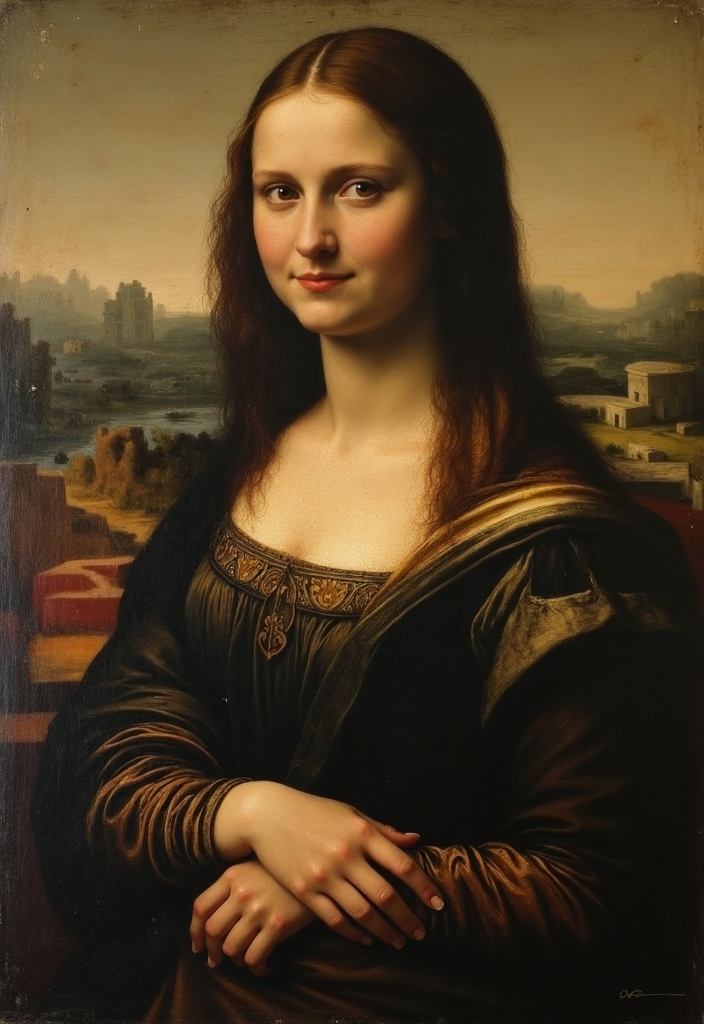}
\end{minipage}

\end{tabular}

\vspace{1mm}

\begin{tikzpicture}[
    optbox/.style={
        draw=dagorange!85!black,
        rounded corners=5pt,
        line width=0.85pt,
        fill=dagyellow!45,
        align=center,
        font=\bfseries\scriptsize,
        inner sep=4pt
    },
    curvearrow/.style={
        -{Latex[length=2.4mm,width=1.8mm]},
        draw=dagorange!90!black,
        line width=1.05pt
    }
]

\node[optbox] (opt) at (0,0) {
Policy Optimization\\[-0.2mm]
{\normalfont\tiny self-reflection + reward update}
};

\draw[curvearrow]
    (-2.05,0.70)
    .. controls (-2.05,-0.10) and ([xshift=-.5cm]opt.west) ..
    ([xshift=-0.05cm]opt.west);

latex\draw[curvearrow]
    ([xshift=0.05cm]opt.east)
    .. controls ([xshift=0.5cm]opt.east) and (2.05,-0.10) ..
    (2.05,0.70);

\end{tikzpicture}

\end{tcolorbox}

}

\end{tabular}
\end{adjustbox}

\caption{
\textbf{Illustration of policy learning.} \textit{ArtMine} converts an evidence-grounded process into a DAG of visual dependencies, compresses the graph into a reconstruction prompt, and refines the generated image through policy learning.
}
\label{fig:process-dag-prompt-final}

\end{figure*}

\subsubsection{Policy Initialization}
The abductive agent ($\mathcal{B}_{\mathrm{abd}}$) is initialized with a policy $\Pi_i = (R, \Sigma, C)$ that governs how it reasons over the evidence. The role $R$ asks $\mathcal{B}_{\mathrm{abd}}$ to act as an art-historical process reconstruction system. Following Peirce's account of abduction as inference to the best explanation \citep{peirce1997pragmatism}, the abductive protocol $\Sigma$ expresses each step as a triplet: an \textbf{observation} of a feature of the work, a \textbf{rule} stating that a process of type $P$ would produce that feature, and a \textbf{hypothesis} that selects the concrete production action $P$ which, under the stated rule, best explains the observed feature. Unlike the original Peircean triplet, which leaves the inferred process underspecified for downstream generation, $\mathcal{B}_{\mathrm{abd}}$ extends $\Sigma$ by adding both an \textbf{action} and an \textbf{evidence} to each inferred step. The action translates the abductive hypothesis into a concrete production instruction that can be carried forward by later modules, such as DAG construction and prompt generation. The evidence then binds the observation of the step to a specific unit of the repository and carries the reliability tag of the unit, so the action remains traceable to its supporting source \citep{howell2001reliable}. The constraint set $C$ further requires the agent to cite evidence, avoid unsupported claims, and remain close to the production logic of the painting. The complete policy is provided in
Figure~\ref{fig:base_policy} and prompt for the agent is provided in the Appendix \ref{app:prompt-templates}.

Under this policy, the agent reconstructs an initial trajectory $\tau = (s_1,
\dots, s_N)$ in production order. It observes the reference work and, at each
step, emits an abductive quadruple grounded in a repository unit and conditioned
on the steps already committed. Since the policy requires each abductive quadruple to include an evidence key if no such key can be supplied, the step is treated as unsupported and excluded from the trajectory. In this way, the trajectory is grounded and provides an initial evidence-linked explanation of how the artwork may have been made.

\subsubsection{Policy Learning}

The process is abduced by the initial policy from $\mathcal{M}_{\text{evidence}}$ alone and is never tested against the way its choices
look once drawn. The initial process is therefore unverified, and it may
misplace an element, misjudge a proportion, or set the wrong palette.

Correcting the initial process requires rendering it as an image and comparing the result with the reference artwork. Prior work on self-evolving agents has shown that feedback from past trajectories can improve an agent’s capabilities through policy optimization \cite{zhang2024agent, chen2025swe, yin2025godel, kakade2026polaris,liu2026escher}. Motivated by this view, we formulate policy learning as multi-reward-driven self-reflection, where reconstruction feedback is used to update the abductive policy from ($\Pi_i$) to ($\Pi_{i+1}$). The abductive agent compares the rendered image with the reference artwork, analyzes deviations in elements such as composition, palette, position, and form, and reflects on the likely causes of these errors. The resulting feedback is incorporated into the policy for the next iteration, so that subsequent trajectories remain consistent with both the evidence repository and the visual appearance of the target artwork. The components of this loop are described below.

\paragraph{Compositional graph construction.}
From $\tau$, the \textbf{composition agent} reads each step and extracts the specification of the corresponding visual element, together with attributes such as colour, size, position, and brushwork texture. For example, a step such as ``paint a seated central figure in muted earth tones with softly folded hands in the lower foreground'' is decomposed into an element node for the sitter, attribute nodes for palette and scale, and relational information specifying the position of the hands. The agent then compiles the trajectory into a directed acyclic graph (DAG) $G$, whose nodes are visual elements and whose edges connect each element to the element it builds on.

Each edge in $G$ has two labels. The first label specifies the relation between the child element and its parent, such as whether the child refines, overlaps, sits over, or accents the parent. The second label specifies the relative position of the child with respect to the parent, for example, whether a hand appears in the lower foreground of the sitter, or a bridge is placed in the distant background behind the figure.
The elements whose \textbf{out-degree} exceeds a predefined threshold are selected as \textbf{focal elements}; that is, they are the visual elements on which many other elements depend. These focal elements jointly anchor the composition. Unlike a flat list of steps, the DAG preserves both the partial order of making and the relational-spatial structure needed for the artwork reconstruction prompt. The prompt for the composition agent is provided in the Appendix \ref{app:dag-construction-prompt}.

\paragraph{Artwork reconstruction prompt.}
The \textbf{visual prompt agent} composes the artwork reconstruction prompt from $G$. It leads with the focal element, then places each remaining element by its relative position to the element it builds on, carrying the attributes from the attribute nodes. It drops artist and title identifiers, so the generator works from the reconstructed appearance rather than from a named work it already knows. A text-to-image model renders the prompt in a generated image $I$. The prompt for the visual prompt agent is provided in the Appendix \ref{app:prompt-construction-prompt}.

\paragraph{Multi-reward scoring.}
We compare $I$ with the reference artwork $I_{\mathrm{ref}}$ along three complementary rewards. The \textbf{style reward} (CSD \cite{somepalli2024measuring}) measures agreement in brushwork, palette, and surface character. The \textbf{perceptual reward} (LPIPS \cite{zhang2018unreasonable}), measures low-level texture and structural agreement. The \textbf{semantic reward} (CLIP \cite{radford2021learning}) measures whether the depicted content matches the reference.

\paragraph{Self-reflection and Policy updation.}

The abduction agent closes the loop by working out why the rendering falls short and what should change. It looks at the rendered and reference artworks together, guided by the multi-reward vector and the ordered actions that produced the image. Reading the rendering against the reference artwork element by element, it notes where an element is missing or where its colour, size, position, or shape has drifted, and assesses the severity of each divergence. What it learns from this falls into two categories. The first, the method-principles $\mathcal{C}_g$, are generic to the art-creation process, such as how to order layers or set a palette from the evidence, so they are folded back into the policy $\Pi_i$ and carried forward to every subsequent run.

The second category consists of artwork-specific corrections $\mathcal{C}_a$, which capture adjustments required by the particular target work, such as softening the sitter's smile, repositioning the folded hands in the lower foreground, refining the atmospheric background, etc. Rather than being converted into general rules, these corrections are stored in a per-work memory $\mathcal{M}_{\mathrm{art}}$. During the process discovery of the next artwork, this memory is provided as a set of worked examples that the agent may adapt when the new evidence supports a similar compositional or material decision.

When the rendered artwork still exhibits substantial divergence from the reference, the reflection $r_\Pi$ produced under the current policy is used to revise the policy. The updated policy is formed as
\begin{equation}
\Pi_{i+1} = (R, \Sigma \cup \mathcal{C}_g, C \cup \mathcal{C}_a),
\end{equation}
where $\mathcal{C}_g$ extends the abductive protocol $\Sigma$ with reusable process-level principles, and $\mathcal{C}_a$ extends the constraint set $C$ with artwork-specific corrections. Duplicate entries are discarded. The loop then reruns under $\Pi_{i+1}$ and continues as long as the rendered artwork improves. An example of an updated policy is presented in the Figure \ref{fig:updated_policy}.
The self-reflection loop stops when one of three conditions is met: \textbf{Acceptance}, where the mismatch severity becomes small, and all rewards stop changing meaningfully; \textbf{No-progress}, where the mean reward declines for $K$ consecutive iterations; or \textbf{Budget}, where the iteration limit is exhausted.
Since a later update can sometimes lower a
reward, the system returns the best-scoring iteration rather than the last. This state consists of the inferred
trajectory $\tau$, the generated image $I$, the reward vector $s$, and a
termination label indicating whether the loop ended by acceptance, no-progress,
or budget exhaustion. 


\subsubsection{Formalizing the learned policy via Cross-Artwork Transfer}

\textit{ArtMine} applies the policy learning loop across a sequence of canonical artworks
$A_1, \dots, A_T$. The learned policy is carried forward, so the method-principles
$\mathcal{C}_g$ and artwork memory $\mathcal{M}_{\mathrm{art}}$ accumulate into a
trained policy $\Pi^\ast$.

After learning across the canonical sequence, \textit{ArtMine} freezes the
policy $\Pi^\ast$ and applies it once to a new artwork, abducing and rendering a
trajectory with no further updates. Any gain must then come from the inherited
method-principles and artwork memory, so this setting tests whether the policy
transfers to a low-evidence regime.

\textit{ArtMine} renders the discovered trajectory in three modes.
\textbf{Single-pass} renders the full trajectory in one pass, as during
learning, yielding one image from the complete prompt.
\textbf{Direct stage-wise} splits the trajectory into ordered stages, each
combining one or more steps of the $\tau$, and renders every stage independently from its
partial DAG. \textbf{Guided stage-wise} renders the stages in sequence,
conditioning each on the previous image so later stages layer over earlier ones. These three modes let us assess how good the discovered process is and whether
it can be rendered step by step.

\section{Experimental setup.}
\begin{figure}[t]
    \centering

    \begin{subfigure}{\columnwidth}
        \centering
        \includegraphics[
            width=\columnwidth,
            keepaspectratio
        ]{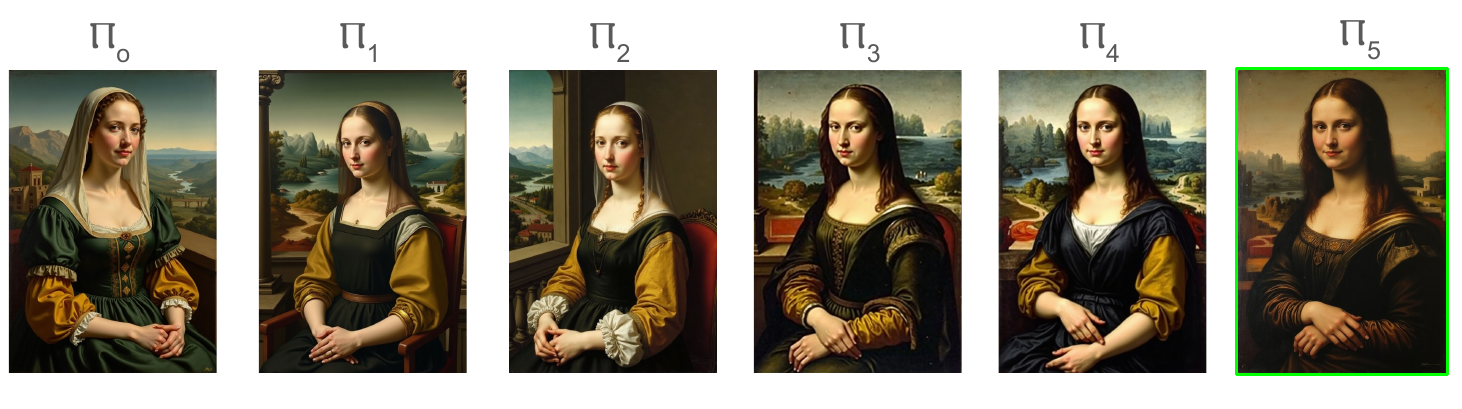}
        \caption{Generated-image trajectory across successive policies $\Pi_0, \Pi_1, \Pi_2, \ldots$.}
        \label{fig:policy_trajectory_a}
    \end{subfigure}

    \vspace{4pt}

    \begin{subfigure}{\columnwidth}
        \centering
        \includegraphics[
            width=\columnwidth,
            keepaspectratio
        ]{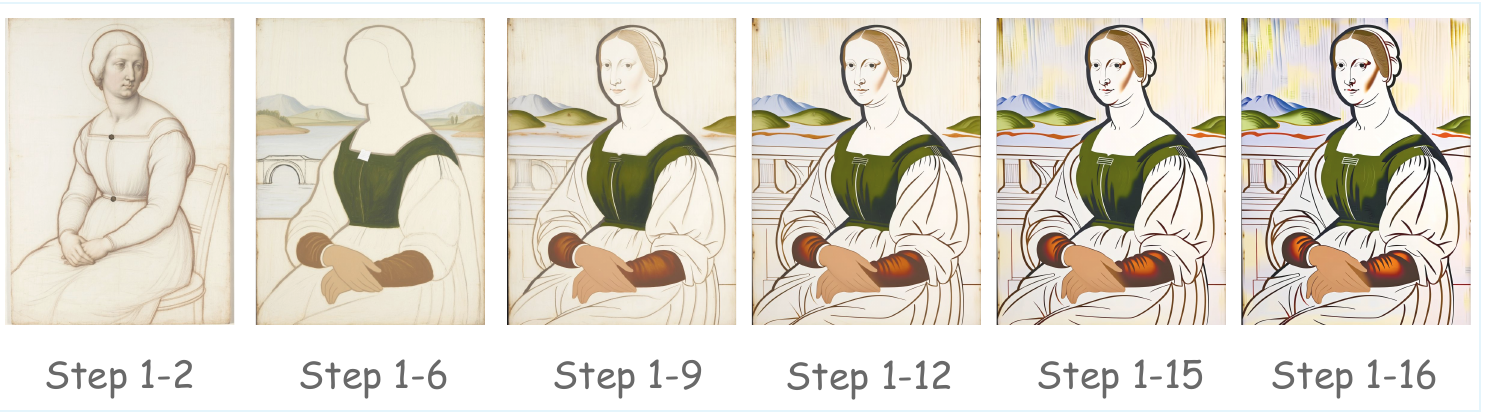}
        \caption{Guided generation highlighting process progression.}
        \label{fig:policy_trajectory_b}
    \end{subfigure}

    \caption{
    \textbf{Process discovery and staged reconstruction.}
    (a) Generated images obtained across successive policy updates, showing how the reconstruction improves as the policy is optimized.
    (b) The highest-scoring process is selected, and its guided stage-wise generation is rendered to visualize the inferred progression from intermediate steps to the final image.
    }
    \label{fig:policy_trajectory}
\end{figure}
\subsection{Experimental details}
Evidence construction is performed with a Deep Research agent built on MiroThinker 1.7 mini \cite{team2025mirothinker}. Given an artwork and artist name, the agent retrieves open-domain evidence and organizes it into the concepts schema described in Section~\ref{sec:evidence}. The remaining reasoning components share the same backbone: the abductive solver, composition agent, and visual prompt generation agent all use Qwen2.5-VL. Image generation is performed with FLUX.1-dev. 
\begin{figure}[t]
    \centering

    \includegraphics[
        width=0.95\columnwidth,
        keepaspectratio
    ]{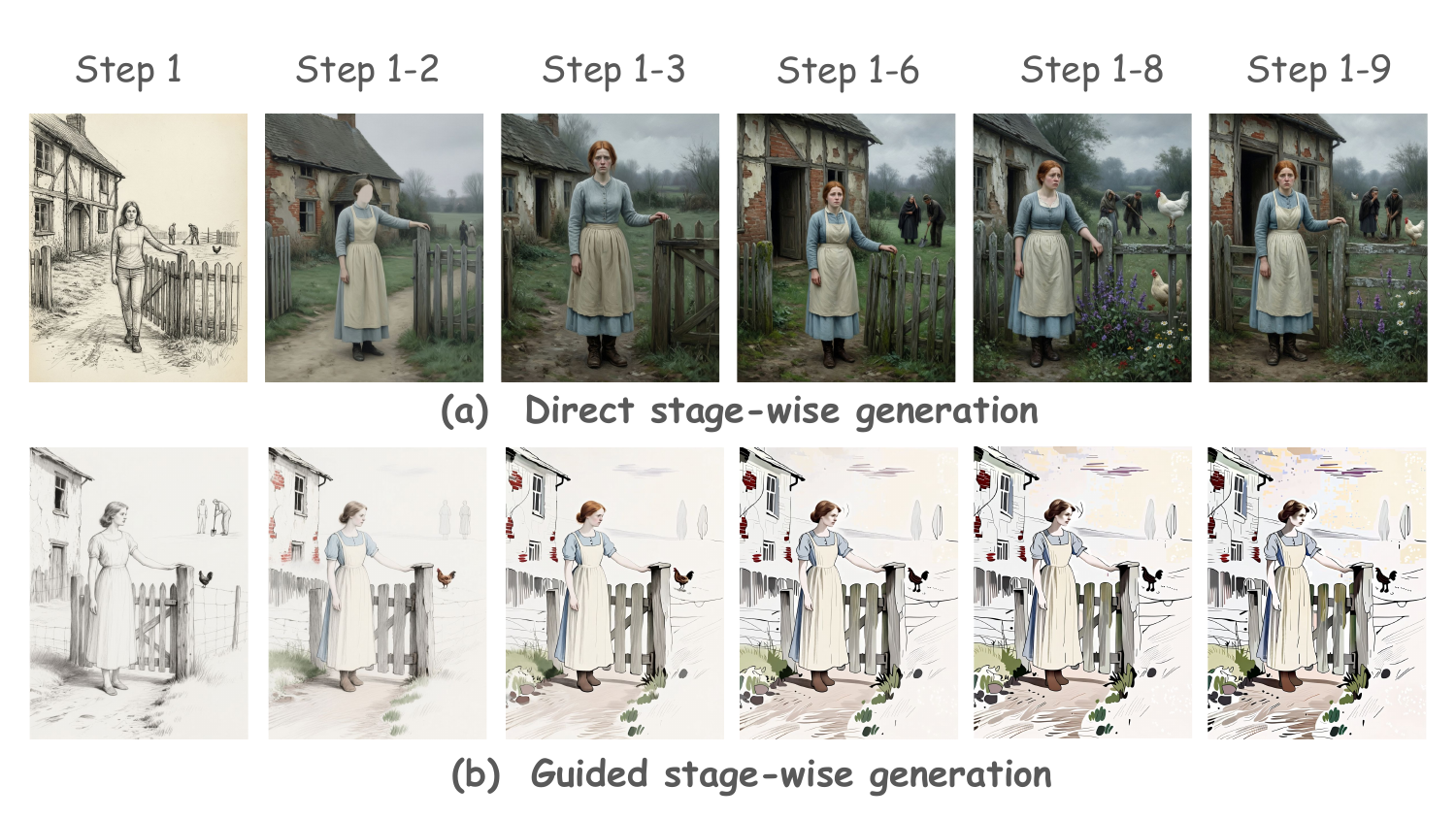}

    \caption{
    \textbf{Direct and guided staged reconstruction for \textit{The Girl at the Gate} by George Clausen.}
    The top panel shows direct stage-wise generation, where each stage is generated independently. The bottom panel shows guided stage-wise generation, where each stage is conditioned on the previous generated stage, resulting in a more coherent reconstruction trajectory.
    }
    \label{fig:trajectories}
\end{figure}
We evaluate process discovery through artwork reconstruction. We compare the generated artwork against the reference image along three complementary axes. CSD measures stylistic similarity, especially brushwork, palette, and surface character. LPIPS measures low-level perceptual distance. CLIP measures semantic agreement between the generated image and the reference content. We track the three scores separately rather than collapsing them into a single scalar, because the weakest axis identifies the main failure mode that reflection should address in next iteration.

\subsection{Dataset}
Following
\cite{kumarartpeer}, we use ten artworks sourced from WikiArt, split into five canonical and five
non-canonical works. The canonical set
comprises \textit{Impression, Sunrise}, \textit{Caf\'e Terrace at Night},
\textit{Guernica}, \textit{The Scream}, and the \textit{Mona Lisa}. The
non-canonical set comprises \textit{The Girl at the Gate} by George Clausen,
\textit{Horse and Cart} by Georges Seurat, \textit{Two Riders by a River} by
Edgar Degas, \textit{Cat and Canary} by Will Barnet, and \textit{A Walk at Dusk}
by Caspar David Friedrich. 

\subsection{Baselines}

We compare the \textit{ArtMine} against a set of prompting strategies that
share its backbone and generator, but omit its policy learning.

\textbf{Prompting strategies.} We compare against four prompting baselines using the same Qwen2.5-VL backbone and FLUX.1-dev generator: CoT, CoT-SC, ToT, and Self-Refine. Each baseline reconstructs the process in a single pass and is evaluated with and without $\mathcal{M}_{\mathrm{evidence}}$. Full prompts are provided in the Appendix.(Refer Figures \ref{fig:prompt_cot}, \ref{fig:prompt_cot_sc}, \ref{fig:prompt_self_refine}, and \ref{fig:prompt_tot})

\textbf{$\mathcal{M}_{\mathrm{evidence}}$ ablation.} We run every strategy in two settings.
\textbf{With $\mathcal{M}_{\mathrm{evidence}}$}, the strategy receives the structured evidence
repository $\mathcal{M}_{\mathrm{evidence}}$ for the artwork, so its reconstruction can be
grounded in the documented materials, technique, and iconography.
\textbf{Without $\mathcal{M}_{\mathrm{evidence}}$}, the same strategy receives only the
reference image and must infer the process from what it observes.

\begin{figure}[t]
    \centering

    \includegraphics[
        width=0.98\columnwidth,
        keepaspectratio
    ]{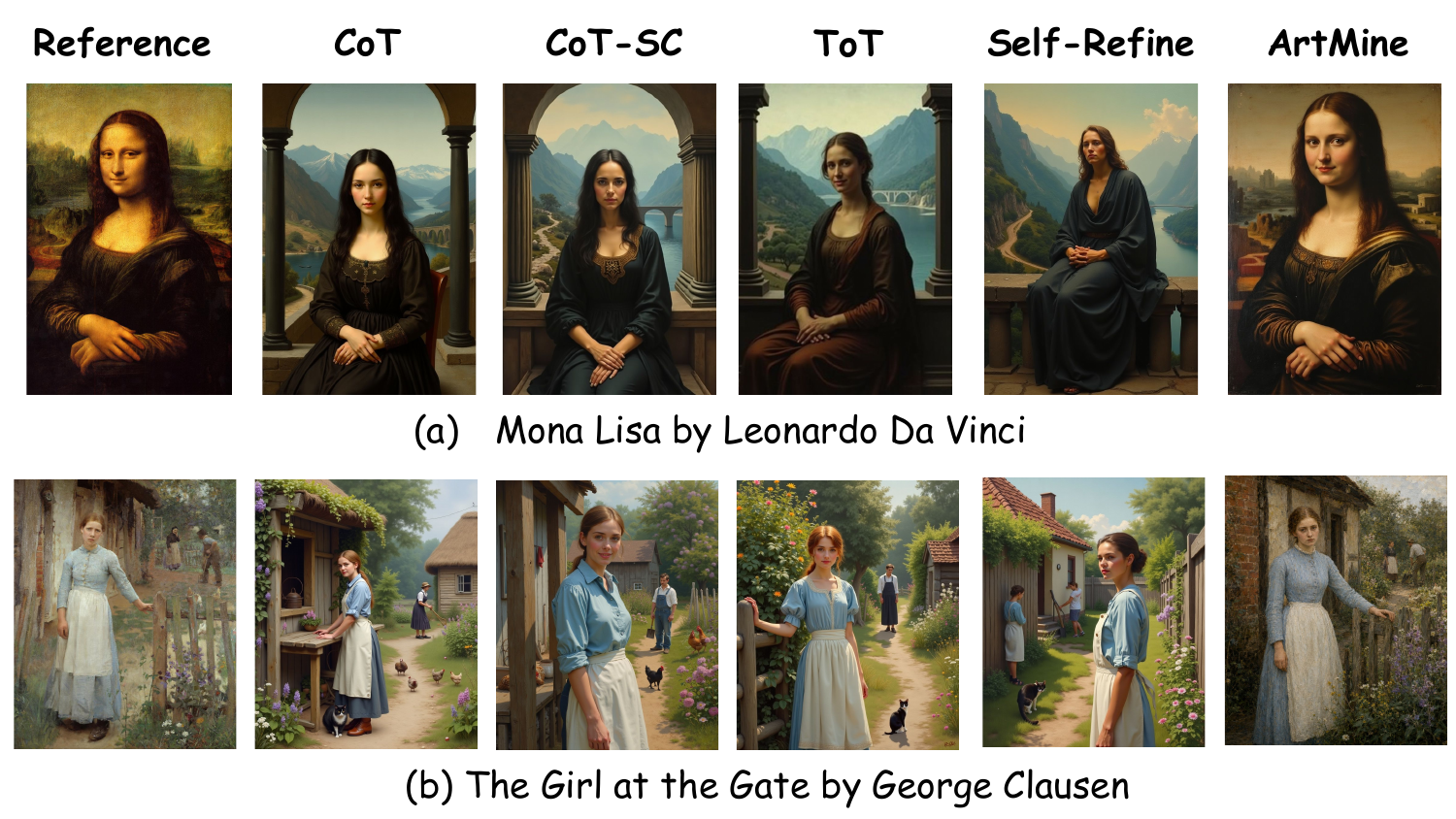}

    \caption{
    Qualitative comparison of reconstruction methods on a canonical and a non-canonical artwork. Each row shows the reference image followed by reconstructions from CoT, CoT-SC, ToT, Self-Refine, and ArtMine. ArtMine better preserves global composition, motif placement, and painterly treatment than the baseline prompting strategies.
    }
    \label{fig:qualitative_comparison_two_rows}
\end{figure}
\section{A Case Study}
\label{sec:experiments}

We illustrate the framework through artwork reconstruction on the canonical and
non-canonical works. The study is intended as a proof-of-concept
exploration of how abductive policy learning and structured evidence synthesis support
interpretable representations of artistic process.


\subsection{Structured Evidence Repository}
\textbf{$\mathcal{M}_{\mathrm{evidence}}$ helps to discover production order. }$\mathcal{M}_{\mathrm{evidence}}$ organizes art-historical information as constraints on how the artwork could have been made. These constraints give the abductive agent the evidence it reasons over to
recover a plausible order of production. For the \textit{Mona Lisa}
(Appendix~\ref{app:memory-monalisa}), evidence about the thin poplar panel
grounds the support that must precede all painting, and infrared findings recover
the chair, balusters, etc., the darkened surface hides. Technical
evidence about sfumato then grounds the later steps that model the mouth, eyes,
hands, and landscape. From these constraints, the
agent infers the production order, placing the support first and the sfumato
modelling last, grounded in documented technique rather than visual appearance.

\textbf{Preserved conflicts guide the abduction agent. }$\mathcal{M}_{\mathrm{evidence}}$ also reduces overconfident reconstructions by
preserving the reliability of each claim. Each item is tagged by evidence type,
so interpretive readings are not weighted like museum or conservation records,
and as Figure~\ref{fig:source-categories} shows, most support comes from those
sources. Conflicts are kept rather than resolved: for \textit{The Scream},
$\mathcal{M}_{\mathrm{evidence}}$ records disagreement over the figure's sex and
dress, which prevents the agent from committing to unsupported details and abduce an androgynous figure. (Refer to Section \ref{app: scream_ev}).

\subsection{Abductive Process Discovery}
\textbf{Visuals alone cannot determine process.} The abductive agent infers a plausible artistic process from the evidence stored in $\mathcal{M}_{\mathrm{evidence}}$. For each observed feature, it tries to hypothesise which operation naturally explains its presence. A visual description of the \textit{Mona Lisa} may note a seated woman, folded hands, a dark costume, and a distant landscape, but not the process behind it. The abduction agent bridges this gap by inferring the sequence of artistic operations that could have produced the observed work.

\textbf{Abduction bridges appearance and process.} The Appendix~\ref{app:trajectory-monalisa} synthesis output illustrates this for the support step. The
observation is a large, thin poplar panel, and the rule is that any support
precedes all paint. From these, the agent infers that Leonardo prepared the
panel first and adds panel preparation as the opening action. Infrared evidence
of a chair, balusters, and a balustrade works the same way. The agent infers
that the pose and loggia setting were planned early, and adds an underdrawing
action before any modelling. Figure~\ref{fig:policy_trajectory} (b) shows these stages in order, from layout
through dead colouring, first modelling, upper layers, synthesis, and a final
surface adjustment. The same pattern holds beyond the \textit{Mona Lisa}. In
\textit{Caf\'e Terrace at Night}, the glowing terrace and blue-violet street
become a plan to organize the cafe, street, and sky first, then build the
warm--cool contrast through colour and brushwork (Refer to Sectionn \ref{app: cafe_process}).

\begin{table}[t]
\centering
\scriptsize
\setlength{\tabcolsep}{4pt}
\renewcommand{\arraystretch}{1.02}
\caption{
Average reconstruction quality across \textbf{canonical artworks}.}
\label{tab:avg_scores}

\resizebox{\linewidth}{!}{%
\begin{tabular}{@{}llccc@{}}
\toprule
\textbf{Setting} & \textbf{Method} & \textbf{CSD $\uparrow$} & \textbf{LPIPS $\downarrow$} & \textbf{CLIP $\uparrow$} \\
\midrule

\multirow{4}{*}{w/o $\mathcal{M}_{\mathrm{evidence}}$}
& CoT & 0.289 & 0.647 & 0.858 \\
& CoT-SC & 0.296 & 0.638 & 0.854 \\
& Self-Refine & 0.268 & 0.634 & 0.855 \\
& ToT & 0.269 & 0.623 & 0.855 \\
\midrule

\multirow{4}{*}{w/ $\mathcal{M}_{\mathrm{evidence}}$}
& CoT & 0.350 & 0.616 & 0.844 \\
& CoT-SC & 0.312 & 0.642 & 0.843 \\
& Self-Refine & 0.291 & 0.625 & 0.840 \\
& ToT & 0.294 & 0.667 & 0.842 \\
\midrule

& \textit{ArtMine} & \textbf{0.395} & \textbf{0.527} & \textbf{0.916} \\

\bottomrule
\end{tabular}%
}
\end{table}

\begin{table}[t]
\centering
\small
\setlength{\tabcolsep}{6pt}
\renewcommand{\arraystretch}{1.12}
\caption{
Average reconstruction quality across the \textbf{non-canonical artworks}. For stage-wise reconstructions, scores are computed only on the final generated stage, since it represents the completed reconstruction.
}
\label{tab:noncanonical_three_avg_scores}

\resizebox{\linewidth}{!}{%
\begin{tabular}{llccc}
\toprule
\textbf{Setting} & \textbf{Method} & \textbf{CSD $\uparrow$} & \textbf{LPIPS $\downarrow$} & \textbf{CLIP $\uparrow$} \\
\midrule

\multirow{4}{*}{Without $\mathcal{M}_{\text{evidence}}$}
& CoT & 0.331 & 0.654 & 0.869 \\
& CoT-SC & 0.264 & 0.654 & 0.853 \\
& Self-Refine & 0.297 & 0.626 & 0.862 \\
& ToT & 0.271 & 0.655 & 0.872 \\

\midrule

\multirow{4}{*}{With $\mathcal{M}_{\text{evidence}}$}
& CoT & 0.347 & 0.601  & 0.873  \\
& CoT-SC & 0.282  & 0.622  & 0.865  \\
& Self-Refine & 0.323 & 0.617 & 0.873 \\
& ToT & 0.289 & 0.593  & 0.882 \\

\midrule

& \textit{ArtMine} (Single-Pass) & \textbf{0.472} & \textbf{0.531} & {0.893}  \\
& \textit{ArtMine} (Stage-Wise (Direct)) & {0.462} & {0.578} & \textbf{0.898}  \\
& \textit{ArtMine} (Stage-Wise (Guided)) & {0.222} & {0.738} & {0.838}  \\

\bottomrule
\end{tabular}%
}
\end{table}

\subsection{Policy Optimization and Transfer to Non-Canonical Artworks}

\textbf{Learned policy tends to improve reconstruction. }The base policy $\Pi_0$ carries only the Peircean structure
(Figure~\ref{fig:base_policy}), which grounds an initial trajectory. Policy learning improves this policy by
comparing the rendered image generated using the initial process with the reference. On the canonical works \textit{ArtMine} outperforms every
prompting baseline, with better CSD, CLIP, and LPIPS
(Table~\ref{tab:avg_scores}; Figures~\ref{fig:qualitative_comparison_two_rows}
and~\ref{fig:qualitative_comparison_seven_artworks}). For non-canonical artwork, where the final learned policy is utilized, \textit{ArtMine} still outperform baselines
(Table~\ref{tab:noncanonical_three_avg_scores};
Figure~\ref{fig:qualitative_comparison_non_canon_artworks}), indicating that it carries the process knowledge across artworks.

\textbf{Rendering the underlying process remains more challenging than rendering the final image. }For non-canonical artworks, image generation is also done stage-wise. Without conditioning on the prior stage, direct stage-wise generation has no partial state to build from, so it renders each stage's prompt as a complete image. As a result, early stages appear complete rather than as intermediate steps in the process. This raises the final-stage fidelity score but does not expose the process's intermediate progression. Guided generation conditions each stage on the previous image, preserving continuity across the sequence. But since this conditioning is image-to-image rather than a fresh render only from the prompt, early errors propagate and compound, causing the final output to drift from the reference (Refer to Table \ref{tab:noncanonical_three_avg_scores} and Figures~\ref{fig:trajectories} and~\ref{fig:noncanonical_direct_guided_stages}). Policy learning and transfer may have improved final artwork reconstruction, but faithful process rendering is still difficult. This leaves two open questions: how a generative system can learn to
represent the artistic process rather than only the appearance of finished artworks,
and whether intermediate artistic states such as underdrawing, blocking-in, and
glazing can be evaluated as meaningful outputs rather than treated only as steps
toward a final image.

\section{Discussion and Limitations}
\label{sec:limitations}

Our work explores a process-centered perspective on computational creativity by shifting attention from the generation of artistic artifacts to the discovery and formalization of creative workflows. Rather than treating artworks as isolated outputs, the proposed framework models artistic production as a sequence of situated decisions grounded in material practice, historical context, and stylistic evolution. This framing aligns with broader directions in human-AI co-creativity that emphasize interaction, interpretation, and reflective collaboration over autonomous artifact generation. By maintaining explicit links between inferred actions and supporting evidence, the framework also moves toward more interpretable and auditable forms of creative AI systems.

A central contribution of the work is the formulation of artistic process inference as an abductive reasoning problem over heterogeneous historical evidence. Unlike domains with formal verification signals, artistic workflows are inherently underdetermined: multiple plausible trajectories may explain the same completed work, and historical documentation is often incomplete, contradictory, or unevenly preserved. Consequently, the framework should not be understood as recovering definitive historical truth, but as constructing evidence-grounded and procedurally coherent hypotheses that support historical interpretation and creative reasoning. In this sense, the system is intended to complement, rather than replace, expert art-historical analysis.

The proposed framework also highlights the emerging role of research agents as knowledge-synthesis systems for computational humanities and creative domains. While recent agentic systems primarily focus on factual retrieval or task execution, our approach uses iterative evidence gathering, structured memory construction, and abductive reasoning to support process-level inference. More broadly, this suggests opportunities for combining large language models, historical archives, and reasoning frameworks in domains where procedural knowledge is implicit and distributed.

At the same time, the work has several important limitations. First, the framework depends heavily on the availability and quality of historical evidence. Artists with extensive archival documentation, technical studies, or correspondence are substantially easier to model than artists whose practices are sparsely documented. As a result, the system may implicitly privilege historically preserved artistic traditions while underrepresenting oral, collaborative, marginalized, or non-Western creative practices where process documentation is limited or differently structured.

Second, abductive inference in creative domains remains fundamentally uncertain. The inferred workflows represent plausible interpretations conditioned on available evidence and modeling assumptions rather than verified historical reconstructions. Although the framework maintains explicit evidence traceability and structured reasoning, these mechanisms do not eliminate the epistemic ambiguity inherent in modeling historical creative processes.

This work should be viewed as a preliminary proof-of-concept rather than a definitive computational model of artistic creation.
Our evaluation remains exploratory, emphasizing interpretability, coherence, and evidential grounding over objective correctness.
Future work can extend this direction through richer multimodal evidence, uncertainty-aware abduction, interactive co-creative systems, and collaboration with art historians, conservators, and artists.
More broadly, we hope this encourages process-centered approaches to computational creativity, artistic reasoning, and human-AI co-creativity.


\bibliography{references}
\bibliographystyle{icml2026}

\newpage
\appendix
\onecolumn

\section{Related Work}
\label{sec:ext_related}

\subsection{Research Agents, Personas, and Creative AI}

Recent research-agent systems extend retrieval-augmented generation with iterative planning, tool use, reflection, and web interaction \citep{lewis2020retrieval,yao2022react,shinn2023reflexion,nakano2021webgpt}. These systems demonstrate that language agents can gather and synthesize information from heterogeneous online sources through multi-step reasoning and interaction. Our work builds on this direction by using a Deep Research agent to gather evidence from multiple sources, synthesize claims across these sources, assign reliability tags to retrieved evidence, and organize the resulting information into a structured evidence repository for downstream abductive reasoning.
Rather than treating retrieved information as unstructured context for generation, the resulting evidence is organized into structured representations that support downstream reasoning about creative processes.

Role-conditioned and persona-conditioned language agents aim to maintain behavioral or stylistic consistency during generation. Early work represented personas through static textual profiles in dialogue systems \citep{zhang2018personalizing,mazare2018training}, while more recent approaches study role-playing and character-conditioned generation in large language models. ArtPeer \citep{kumarartpeer} explores artist-centered persona construction from art-historical sources, while CharacterGLM \citep{zhou2024characterglm} studies character-consistent generation under persona conditioning. In most of these systems, personas primarily function as conditioning mechanisms for generating stylistically or behaviorally consistent outputs. In contrast, our framework evaluates discovered process trajectories through artwork-grounded policy optimization, where rendered artwork are compared with reference artworks and the resulting deviations drive self-reflection-based policy updates.

Our work is also related to computational creativity and generative art systems. Classical systems such as AARON \citep{cohen1995further} and The Painting Fool \citep{colton2012painting} generate artworks through procedural rules and heuristic creative strategies. More recent generative models such as DALL$\cdot$E~3 \citep{betker2023improving}, Stable Diffusion \citep{rombach2022high}, and Jukebox \citep{dhariwal2020jukebox} learn distributions over artistic artifacts from large-scale multimodal data. Agent-based simulations of social behavior and interaction \citep{park2023generative} similarly model emergent behaviors through multi-agent interaction. In contrast, our work focuses on discovering historically grounded creative trajectories from documentary, archival, technical, and art-historical evidence, and refining them through artwork-grounded self-reflection rather than treating generation as a one-shot artifact-production task.

\subsection{Abductive Reasoning and Process Inference}

Abductive reasoning, commonly described as inference to the best explanation, was introduced by Peirce and later formalized in AI as reasoning over hypotheses that explain observations under background knowledge \citep{peirce1997pragmatism,pople1973mechanization,josephson1996abductive}. Subsequent work has explored logic-based, probabilistic, and neural-symbolic formulations of abduction \citep{kakas1992abductive,ng1992abductive,qiao2023reasoning}. More recently, large language models have demonstrated abductive capabilities in commonsense and explanatory reasoning tasks \citep{bhagavatula2019abductive,weng2023large}. However, these approaches typically rely on unconstrained generation and often lack explicit mechanisms for grounding hypotheses in structured external evidence or comparing competing explanations under domain-specific constraints.

Our framework treats artistic process reconstruction as an abductive inference problem grounded in heterogeneous historical evidence. Artwork features act as observations, documentary and conservation records provide contextual knowledge, and reconstructed production steps form candidate hypotheses. The reasoning process is guided by a structured Peircean abductive framework that maintains explicit connections between observations, inferred actions, and supporting evidence.

The work is also related to procedural and process-oriented modeling systems such as Voyager \citep{wang2023voyager}, ChemCrow \citep{bran2023chemcrow}, and BioPlanner \citep{o2023bioplanner}, which generate executable procedures from unstructured inputs. These approaches often rely on external environments or formal verification signals for evaluation. In contrast, artistic process reconstruction rarely admits definitive correctness criteria because historical workflows are only partially observable. Our framework therefore evaluates reconstructed processes through evidence-grounded plausibility assessments informed by artist-specific and movement-specific context.

\subsection{Computational Art History and Process Discovery}

Prior work in computational art history has focused primarily on analyzing completed artistic artifacts through tasks such as attribution \citep{elgammal2018shape,joshi2026dota, van2015toward, ji2021discerning,ugail2023deep}, classification \citep{karayev2013recognizing,dangeti2024style,saleh2015large,strafforello2025have,limpijankit2026does}, stylistic similarity \citep{seguin2016visual}, interpretability \citep{gatys2015neural,bau2019seeing,schneider2026explainability,bau2018gan}, and style transfer \citep{zhu2017unpaired,karras2020analyzing,he2026freestyle,guo2026artwork}. These approaches have significantly advanced the understanding of visual representations, stylistic structures, and semantic organization in artworks, enabling large-scale analysis of artistic movements, painterly styles, and cross-modal aesthetic relationships. More recently, generative models have further expanded the scope of computational art analysis by learning high-dimensional distributions over artistic artifacts from large multimodal datasets. However, despite these advances, most existing approaches remain artifact-centric: they analyze completed works or synthesize stylistically plausible outputs without explicitly modeling the procedural, historical, and contextual dynamics underlying artistic production.

Our work instead focuses on artistic processes as the primary object of inference. In contrast to tasks such as attribution or style transfer, which operate on observable visual properties of final artworks, artistic process discovery requires reasoning about latent creative trajectories that are only partially reflected in the completed artifact. These trajectories often involve iterative experimentation, revisions, material constraints, environmental influences, and interactions with broader artistic movements or social contexts. Such information is rarely available in a single structured source and instead exists in fragmented forms distributed across archival records, correspondence, technical analyses, interviews, biographies, conservation studies, preparatory sketches, and art-historical scholarship.

Our work is additionally related to process discovery methods developed in areas such as workflow mining and business process analysis \citep{huser2012process,leemans2013discovering}, where latent procedures are reconstructed from structured event logs and execution traces. In these domains, process discovery typically assumes explicit temporal records, well-defined state transitions, and relatively objective notions of correctness. Artistic domains differ substantially because creative workflows are inherently ambiguous, non-linear, and historically underdetermined. Process traces are fragmented, selectively preserved, and frequently mediated through interpretation rather than direct observation. As a result, reconstructing artistic workflows cannot rely on exact procedural recovery and instead requires reasoning under uncertainty over heterogeneous evidence sources.



\section{Populated Memory}
This appendix presents the full memory $M_{\text{evidence}}$ produced by the
deep-research agent for Leonardo da Vinci's \emph{Mona Lisa} and Edvard Munch's \emph{The Scream} , organized across
the eleven typed categories defined in \Cref{sec:evidence}. Each category is
presented as a JSON document with \texttt{evidence\_type} tags, source
citations, and conflict flags as specified by the schema.

\subsection{Mona Lisa}
\label{app:memory-monalisa}

\subsubsection{Metadata}

\begin{footnotesize}
\begin{verbatim}
{
  "metadata": {
    "title": {
      "primary": "Portrait of Lisa Gherardini, wife of Francesco del
                  Giocondo, known as La Joconde or Mona Lisa",
      "alternates": ["Mona Lisa", "La Joconde", "Monna Lisa",
                     "La Gioconda"],
      "evidence_type": "direct_evidence",
      "source": "Louvre Collections, INV 779"
    },
    "artist": {
      "full_name": "Leonardo di ser Piero da Vinci",
      "nationality": "Italian (Florentine Republic)",
      "birth_year": 1452, "death_year": 1519,
      "evidence_type": "direct_evidence"
    },
    "date": {
      "value": "1503-1519",
      "precision": "begun c. 1503; reworked until Leonardo's death in 1519",
      "conflict_flag": true,
      "evidence_type": "direct_evidence",
      "source": "Louvre Collections, INV 779"
    },
    "medium": {
      "value": "Oil on poplar wood panel",
      "evidence_type": "direct_evidence", "source": "Louvre"
    },
    "dimensions": {
      "height_cm": 79.4, "width_cm": 53.4,
      "height_in": 31.3, "width_in": 21.0,
      "orientation": "portrait",
      "evidence_type": "direct_evidence",
      "source": "Louvre"
    },
    "current_location": {
      "institution": "Musee du Louvre, Departement des Peintures",
      "city": "Paris", "wing": "Denon", "level": 1,
      "gallery": "Salle 711 (Salle des Etats)",
      "display": "Climate-controlled protective case behind glass",
      "evidence_type": "direct_evidence",
      "source": "Louvre"
    },
    "identifiers": {
      "inventory_number": "INV 779",
      "other_inventory_number": "MR 316",
      "credit_line": "French royal collection; Louvre collection",
      "canonical_object": "The Louvre painting, not the Prado copy,
                           Isleworth version, or later replicas",
      "evidence_type": "direct_evidence"
    }
  }
}
\end{verbatim}
\end{footnotesize}

\subsubsection{Historical Context}

\begin{footnotesize}
\begin{verbatim}
{
  "historical_context": {
    "origin": {
      "description": "Begun around 1503, probably in Florence, and kept
                      by Leonardo rather than delivered to the Giocondo
                      family",
      "evidence_type": "direct_evidence", "source": "Louvre, INV 779"
    },
    "sitter_identification": {
      "name": "Lisa Gherardini, wife of Francesco del Giocondo",
      "support": "Sixteenth-century testimony and the Agostino Vespucci
                  annotation of October 1503 naming the head of Lisa
                  del Giocondo",
      "evidence_type": "direct_evidence",
      "source": "Louvre; Vespucci annotation, Heidelberg University Library"
    },
    "move_to_france": {
      "description": "Leonardo brought major paintings to France after entering the service of Francis I in 1516; the Mona Lisa was probably acquired by the king in
      1518, likely through paintings sold by Salai",
      "evidence_type": "direct_evidence",
      "source": "Louvre; Jestaz archival discovery, 1518 payment to Salai"
    },
    "working_method": {
      "description": "Late Florentine and French period in which
                      Leonardo worked slowly and revised paintings over
                      long periods, leaving some works incomplete",
      "evidence_type": "interpretation", "source": "Louvre commentary"
    }
  }
}
\end{verbatim}
\end{footnotesize}

\subsubsection{Materials and Technique}

\begin{footnotesize}
\begin{verbatim}
{
  "materials_and_technique": {
    "support": {
      "panel": "Thin poplar wood panel, a support common in Florence",
      "scale": "Large enough to show the sitter at close to natural scale",
      "evidence_type": "direct_evidence", "source": "Louvre"
    },
    "ground": {
      "note": "Support necessarily precedes all paint layers; exact ground preparation is not specified by the Louvre source",
      "evidence_type": "indirect_evidence"
    },
    "costume_layers": {
      "description": "Infrared reflectography clarifies a likely dark
                      green dress with yellow detachable sleeves, a white
                      chemise at the shoulder opening, and a transparent
                      silk veil fixed at the chest with gold-thread
                      embroidery",
      "evidence_type": "direct_evidence", "source": "Louvre, IR commentary"
    },
    "hair_and_veil": {
      "description": "A transparent veil descends over the shoulders;
                      infrared shows the hair held back in a bun or cap
                      with only some strands beside the face",
      "evidence_type": "direct_evidence", "source": "Louvre, IR commentary"
    },
    "sfumato": {
      "description": "Forms modeled through subtle tonal transitions
                      rather than hard contour, especially the face,
                      hands, mouth, and landscape recession",
      "note": "Exact number of layers and sittings is not stated",
      "evidence_type": "interpretation", "source": "Louvre commentary"
    }
  }
}
\end{verbatim}
\end{footnotesize}

\subsubsection{Formal Analysis}

\begin{footnotesize}
\begin{verbatim}
{
  "formal_analysis": {
    "form": {
      "content": "A calm, monumental half-length figure; the body forms
                  a stable pyramidal mass with folded hands as a broad
                  base, softened by sfumato rather than sharp outline",
      "evidence_type": "indirect_evidence", "source": "Louvre"
    },
    "composition": {
      "content": "A centered half-length portrait before a distant
                  landscape; the sitter faces the viewer at a slight
                  angle with folded hands in the lower foreground",
      "evidence_type": "indirect_evidence", "source": "Louvre; IR commentary"
    },
    "color": {
      "content": "Muted, warm-brown palette due to aged varnish; subdued
                  flesh, dark green-brown clothing, transparent veil, and
                  greened landscape tones; non-original varnish has
                  yellowed and darkened the original palette",
      "evidence_type": "direct_evidence", "source": "Louvre, conservation"
    },
    "light": {
      "content": "Soft light over face, chest, and hands without strong
                  cast shadows; gradual modeling and atmospheric, not
                  dramatic, illumination",
      "evidence_type": "indirect_evidence", "source": "Louvre"
    },
    "perspective_and_space": {
      "content": "A near balcony-like interior with a shallow stable
                  foreground opening onto deep landscape recession;
                  infrared clarifies the balustrade, balusters, a
                  rounded-backed chair, and small side columnettes",
      "evidence_type": "direct_evidence", "source": "Louvre, IR commentary"
    }
  }
}
\end{verbatim}
\end{footnotesize}

\subsubsection{Iconographic Analysis}

\begin{footnotesize}
\begin{verbatim}
{
  "iconographic_analysis": [
    {"object": "Seated woman",
     "identity": "Lisa Gherardini, per the Louvre-supported identification",
     "role": "Primary focal subject; central half-length seated figure",
     "evidence_type": "direct_evidence"},
    {"object": "Smile and mouth",
     "construction": "Softly modeled with blurred transitions at the
                      corners, avoiding a hard outline",
     "meaning": "Source of the portrait's psychological ambiguity",
     "evidence_type": "interpretation"},
    {"object": "Eyes and gaze",
     "construction": "Outward gaze framed by soft shadows, hair, and veil",
     "meaning": "Direct engagement with the viewer while preserving reserve",
     "evidence_type": "interpretation"},
    {"object": "Folded hands",
     "construction": "Right hand over the left across the lower foreground",
     "meaning": "Composure and restraint; anchors the lower composition",
     "evidence_type": "indirect_evidence"},
    {"object": "Transparent veil",
     "construction": "Fine veil over head and shoulders; a further
                      transparent silk layer over the dress",
     "meaning": "Refined Florentine costume; pregnancy reading debated",
     "conflict_flag": true,
     "evidence_type": "direct_evidence"},
    {"object": "Dark dress with sleeves",
     "construction": "Likely dark green dress, yellow detachable sleeves,
                      white chemise at the shoulder opening",
     "meaning": "Fashionable dress fit for the wife of a silk merchant",
     "evidence_type": "direct_evidence"},
    {"object": "Chair and balusters",
     "construction": "Rounded-backed 'a pozzetto' chair with balusters,
                      visible in infrared",
     "meaning": "Domestic or loggia-like setting supporting the pose",
     "evidence_type": "direct_evidence"},
    {"object": "Balustrade and side columnettes",
     "construction": "Low barrier with rectangular mouldings and small
                      framing columnettes at the side edges",
     "meaning": "Threshold between the sitter and the vast landscape",
     "evidence_type": "direct_evidence"},
    {"object": "Distant landscape",
     "construction": "Mountains, watercourses, a winding road on the
                      left, and a bridge over a river on the right",
     "meaning": "Idealized landscape expanding the portrait beyond
                 a simple interior likeness",
     "evidence_type": "direct_evidence"},
    {"object": "Absence of jewelry",
     "construction": "No prominent jewels; costume and veil emphasized",
     "meaning": "Read as sobriety, though varnish darkening partly
                 creates this impression",
     "evidence_type": "interpretation"}
  ]
}
\end{verbatim}
\end{footnotesize}

\subsubsection{Documentary Evidence}

\begin{footnotesize}
\begin{verbatim}
{
  "documentary_evidence": {
    "sources": [
      {"source": "Agostino Vespucci marginal annotation",
       "date": "October 1503",
       "location": "Heidelberg University Library",
       "significance": "Compares Leonardo to Apelles and mentions the
                        head of Lisa del Giocondo, confirming the
                        portrait was underway by that date",
       "evidence_type": "direct_evidence"},
      {"source": "Giorgio Vasari, Lives",
       "date": "1550",
       "significance": "States Leonardo undertook a portrait of Mona
                        Lisa, wife of Francesco del Giocondo, and left
                        it unfinished after years of work",
       "editorial_note": "Treated by the Louvre as mediated by
                          informants rather than direct viewing",
       "conflict_flag": true,
       "evidence_type": "direct_evidence"},
      {"source": "Antonio de Beatis, travel diary",
       "date": "10 October 1517",
       "significance": "Records a portrait of a Florentine lady made
                        from life at the request of Giuliano de' Medici,
                        fueling debate over the commission",
       "conflict_flag": true,
       "evidence_type": "direct_evidence"},
      {"source": "Archival payment to Salai (Jestaz)",
       "date": "1518",
       "significance": "A large payment by Francis I to Salai for
                        paintings, used to explain the likely royal
                        acquisition of the Mona Lisa",
       "evidence_type": "direct_evidence"}
    ]
  }
}
\end{verbatim}
\end{footnotesize}

\subsubsection{Provenance}

\begin{footnotesize}
\begin{verbatim}
{
  "provenance": {
    "chain_of_ownership": [
      {"period": "c. 1503 - 1519",
       "owner": "Leonardo da Vinci",
       "transaction": "Begun probably for Francesco del Giocondo but
                       retained by the artist; carried to France in 1516",
       "evidence_type": "direct_evidence"},
      {"period": "c. 1518",
       "owner": "King Francis I of France",
       "transaction": "Probably acquired through the paintings sold by
                       Salai to the king",
       "evidence_type": "direct_evidence"},
      {"period": "16th-18th century",
       "owner": "French royal collection",
       "evidence_type": "direct_evidence"},
      {"period": "present",
       "owner": "Musee du Louvre",
       "transaction": "Held as INV 779",
       "evidence_type": "direct_evidence"}
    ],
    "notes": [
      {"note": "Stolen from the Louvre in 1911 and recovered in Florence
                in 1913, then returned to the Louvre",
       "evidence_type": "direct_evidence"}
    ]
  }
}
\end{verbatim}
\end{footnotesize}

\subsubsection{Scientific Analysis}

\begin{footnotesize}
\begin{verbatim}
{
  "scientific_analysis": {
    "infrared_reflectography": {
      "finding": "Makes visible the rounded-backed chair, balusters,
                  balustrade, side columnettes, and costume and veil
                  details obscured in the present surface",
      "evidence_type": "direct_evidence", "source": "Louvre laboratory"
    },
    "support_condition": {
      "finding": "Thin poplar support, humidity-sensitive, slightly
                  convex, with an 11 cm crack from the top edge through
                  the hair toward the forehead; reverse stabilized with
                  butterfly inserts and pieces of cloth; not cut down",
      "evidence_type": "direct_evidence", "source": "Louvre conservation"
    },
    "varnish": {
      "finding": "Thick, irregular, oxidized non-original varnishes
                  postdating Leonardo yellow and darken the image and
                  shift the blue sky toward green",
      "evidence_type": "direct_evidence", "source": "Louvre conservation"
    },
    "prado_copy_comparison": {
      "finding": "The restored Prado copy clarifies costume and color
                  relationships now hard to see in the Louvre painting",
      "evidence_type": "indirect_evidence",
      "source": "Louvre; Museo Nacional del Prado"
    },
    "contested_imaging_claims": {
      "conflict_flag": true,
      "finding": "Pascal Cotte's claims of hidden earlier portraits,
                  spolvero transfer, or extensive reworking lack broad
                  institutional consensus and are treated cautiously",
      "evidence_type": "speculative"
    }
  }
}
\end{verbatim}
\end{footnotesize}

\subsubsection{Interpretations}

\begin{footnotesize}
\begin{verbatim}
{
  "interpretations": {
    "frameworks": [
      {"id": "renaissance_landmark",
       "claim": "A landmark of High Renaissance portraiture uniting
                 naturalistic likeness, psychological ambiguity,
                 sfumato, and atmospheric landscape",
       "evidence_type": "interpretation"},
      {"id": "ambiguous_smile",
       "claim": "The smile reads as ambiguous because soft modeling
                 around the mouth and eyes avoids a fixed expression",
       "evidence_type": "interpretation"},
      {"id": "mourning_dress",
       "claim": "Darkened clothing read as mourning, though the Louvre
                 attributes the sobriety partly to aged varnish",
       "conflict_flag": true,
       "evidence_type": "interpretation"},
      {"id": "pregnancy_veil",
       "claim": "The transparent veil linked to pregnancy or a
                 guarnello garment, but the figure lacks the rounded
                 form seen in comparison portraits",
       "conflict_flag": true,
       "evidence_type": "interpretation"}
    ]
  }
}
\end{verbatim}
\end{footnotesize}

\subsubsection{Cultural Impact}

\begin{footnotesize}
\begin{verbatim}
{
  "cultural_impact": {
    "recognition": {
      "claim": "Among the most widely recognized paintings in the world
                and central to the public identity of the Louvre",
      "evidence_type": "interpretation"
    },
    "theft": {
      "claim": "The 1911 theft and 1913 recovery greatly increased the
                painting's modern fame and iconic status",
      "evidence_type": "direct_evidence"
    },
    "display": {
      "claim": "Shown in a climate-controlled protective case behind
                glass, reflecting its importance and material fragility",
      "evidence_type": "direct_evidence"
    }
  }
}
\end{verbatim}
\end{footnotesize}

\subsubsection{Conflicts}

\begin{footnotesize}
\begin{verbatim}
"conflicts": [
    {
      "content": "The commission and sitter identification have 
      been debated. Vasari and the Vespucci annotation support Lisa 
      del Giocondo, while Antonio de Beatis records a portrait of a 
      Florentine lady linked to Giuliano de' Medici. The Louvre 
      concludes that the painting represents Lisa del Giocondo, 
      while acknowledging the complexity of the Medici reference.",
      "evidence_type": "direct_evidence",
      "sources": [
        "Louvre Collections, identity and documentary commentary for INV 779",
        "Giorgio Vasari, Lives, 1550",
        "Antonio de Beatis travel diary, 1517"
      ]
    },
    {
      "content": "The meaning of the transparent veil is debated. 
      Some interpretations connect it to pregnancy dress, but the 
      Louvre states that this remains uncertain and notes that the 
      figure lacks the rounded abdomen seen in comparison 
      examples.",
      "evidence_type": "interpretation",
      "sources": [
        "Louvre Collections, costume commentary for INV 779"
      ]
    },
    {
      "content": "Claims of hidden earlier portraits or extensive 
      reworking based on private technical imaging have not reached 
      broad institutional consensus and should not be treated as 
      settled evidence for the painting's construction sequence.",
      "evidence_type": "speculative",
      "sources": [
        "Pascal Cotte publications and public reports on technical 
        imaging claims",
        "Louvre Collections, institutional conservation and 
        examination commentary for INV 779"
      ]
    }
  ],
}
\end{verbatim}
\end{footnotesize}

\subsection{The Scream}
\label{app: scream_ev}
\subsubsection{Metadata}

\begin{footnotesize}
\begin{verbatim}
{
  "metadata": {
    "title": {
      "primary": "The Scream",
      "alternates": ["Skrik", "Verzweiflung", "Despair"],
      "evidence_type": "direct_evidence",
      "source": "Nasjonalmuseet (Oslo) collection catalogue, NG.M.00939"
    },
    "artist": {
      "full_name": "Edvard Munch",
      "evidence_type": "direct_evidence",
      "source": "Nasjonalmuseet (Oslo) collection catalogue, NG.M.00939"
    },
    "date": {
      "value": "1893",
      "precision": "painted 1893; inscription added c. 1904",
      "evidence_type": "direct_evidence",
      "source": "Nasjonalmuseet (Oslo) collection catalogue, NG.M.00939"
    },
    "medium": {
      "value": "Tempera, oil, pastel (crayon) on cardboard",
      "evidence_type": "direct_evidence",
      "source": "Nasjonalmuseet (Oslo) collection catalogue, NG.M.00939"
    },
    "dimensions": {
      "height_cm": null, "width_cm": null,
      "dimensions_note": "91 x 73.5 cm",
      "orientation": "portrait",
      "evidence_type": "direct_evidence", "source": "Nasjonalmuseet (Oslo)"
    },
    "current_location": {
      "institution": "Nasjonalmuseet for kunst, arkitektur og design (National Museum)",
      "city": "Oslo, Norway",
      "gallery": "Room 060, The Collection Exhibition - Edvard Munch",
      "evidence_type": "direct_evidence", "source": "Nasjonalmuseet (Oslo)"
    },
    "identifiers": {
      "accession_number": "NG.M.00939",
      "credit_line": "Gift from Olaf Schou, 1910",
      "catalogue_numbers": ["Woll M 333"],
      "canonical_object": "The 1893 National Museum (Oslo) painted version, not the
                           1910 Munch Museum version or the pastel versions",
      "inscription": "Signed and dated lower left: E. Munch 1893",
      "evidence_type": "direct_evidence"
    }
  }
}
\end{verbatim}
\end{footnotesize}

\subsubsection{Historical Context}

\begin{footnotesize}
\begin{verbatim}
{
  "historical_context": {
    "origin": {
      "description": "Painted in 1893, as confirmed by the museum catalogue",
      "evidence_type": "direct_evidence",
      "source": "Nasjonalmuseet (Oslo) collection catalogue, NG.M.00939"
    },
    "scene_description": {
      "description": "Depicts a solitary figure against the Oslofjord at sunset,
                      near Ekeberg. Munch's 1892 prose poem describes the sky
                      turning blood-red above the blue-black fjord while he
                      stood trembling with anxiety and sensed an infinite
                      scream passing through nature",
      "evidence_type": "direct_evidence",
      "source": "Munch's own notebooks and correspondence, quoted in museum catalogues"
    },
    "stylistic_shift": {
      "description": "Marks Munch's shift from Symbolism toward Expressionism,
                      fusing form and content so line and color carry meaning",
      "evidence_type": "interpretation",
      "source": "Published scholarship on The Scream and museum exhibits"
    }
  }
}
\end{verbatim}
\end{footnotesize}

\subsubsection{Materials and Technique}

\begin{footnotesize}
\begin{verbatim}
{
  "materials_and_technique": {
    "support": {
      "panel": "Cardboard, a cheap support Munch used extensively 
      early in his career; conservation analysis notes the board is 
      of poor quality",
      "evidence_type": "direct_evidence", "source": "Nasjonalmuseet 
      (Oslo); Pettersen et al., 2022"
    },
    "execution_style": {
      "description": "Quick, spontaneous technique using tempera and
        oil-based pigments plus pastel/crayon; the lack of fine 
        detail and the fluid strokes suggest rapid execution",
      "evidence_type": "direct_evidence", 
      "source": "Nasjonalmuseet (Oslo) collection catalogue"
    },
    "layer_order": {
      "description": "Sky and water likely laid in first as broad washes
                      (cadmium yellow and vermilion for the sky, ultramarine
                      and green for the water), then the walkway and railing
                      in ochre and brown, followed by the two dark-clothed
                      background figures",
      "note": "Layer order is inferred from visual and conservation
               evidence, not a documented stratigraphy",
      "evidence_type": "indirect_evidence", "source": "Pettersen et al., 2022"
    },
    "figure_details": {
      "description": "Foreground figure's head filled with flat light-ochre
                      ground outlined in white pastel, with mouth, nostrils,
                      and eyes added in dark crayon, giving a drawing-like
                      look; the painting is signed twice in pencil",
      "evidence_type": "indirect_evidence",
      "source": "Pettersen et al., 2022; Munchmuseet conservation research"
    },
    "pigments": {
      "description": "Pigment analysis shows 19th-century paints, notably
                      cadmium yellow, vermilion, ultramarine, and viridian,
                      along with organic dyes; cadmium yellow on cardboard
                      contributes to the painting's long-term fragility",
      "evidence_type": "direct_evidence", 
      "source": "Munchmuseet conservation research; Pettersen et al., 2022"
    }
  }
}
\end{verbatim}
\end{footnotesize}

\subsubsection{Formal Analysis}

\begin{footnotesize}
\begin{verbatim}
{
  "formal_analysis": {
    "form": {
      "content": "Simplified, schematic forms; the anguished figure is
                  painted with flat, loose strokes, a skull-like head as a
                  pale ochre oval outlined in white, a wide black mouth,
                  dot-eyes, and a dark flowing body of indistinct clothing",
      "evidence_type": "indirect_evidence", "source": "Supplied image; Nasjonalmuseet (Oslo)"
    },
    "composition": {
      "content": "A dynamic diagonal composition; the boardwalk's receding
                  orange-brown planks lead from the lower right toward a
                  horizontal band of bright orange sky, two small blue-clad
                  figures walk away on the left, and the foreground is
                  dominated by a single androgynous figure clutching its face",
      "evidence_type": "indirect_evidence", "source": "Supplied image"
    },
    "color": {
      "content": "Vibrant, non-naturalistic colors dominate; blood-red and
                  orange in the sky set in high contrast against a deep
                  blue-black fjord, largely unblended to heighten emotion",
      "evidence_type": "direct_evidence", "source": "Munch's own notebooks; supplied image"
    },
    "light": {
      "content": "The sky is painted in tongues of fire, red and orange,
                  above a deep blue-black fjord, matching the sunset
                  described in Munch's prose poem, creating strong tonal
                  tension",
      "evidence_type": "direct_evidence", "source": "Munch's own notebooks; supplied image"
    },
    "perspective_and_space": {
      "content": "Strong linear perspective, with the converging railing
                  and road pulling the eye toward the horizon; the
                  screaming figure holds the foreground while the two
                  friends recede along the left path and the fjord and
                  hills lie beyond, emphasizing the figure's isolation",
      "evidence_type": "indirect_evidence", "source": "Supplied image; Munch's own notebooks"
    }
  }
}
\end{verbatim}
\end{footnotesize}

\subsubsection{Iconographic Analysis}

\begin{footnotesize}
\begin{verbatim}
{
  "iconographic_analysis": [
    {"object": "Screaming figure",
     "identity": "Central androgynous figure clutching its face",
     "construction": "Dominates the foreground on the boardwalk, hands
                      framing the face, mouth agape, skull-like head with
                      hollowed eyes; pale ochre head outlined in white,
                      wide dark mouth and eyes, dark flowing body",
     "meaning": "Read as a universal figure expressing existential anxiety
                and alienation; identity deliberately ambiguous in gender
                and dress",
     "evidence_type": "interpretation"},
    {"object": "Red-orange sky",
     "construction": "Broad swirling horizontal bands above the horizon,
                      tongues of fire in red and orange, curving to echo
                      the figure's silhouette",
     "meaning": "Externalizes the anxiety described in Munch's poem,
                linking human emotion to nature",
     "evidence_type": "direct_evidence"},
    {"object": "Fjord",
     "construction": "Cool bands of blue-black and blue-green beneath the
                      fiery sky",
     "meaning": "Still, cool counterpart to the agitated sky",
     "evidence_type": "direct_evidence"},
    {"object": "Boardwalk and railing",
     "construction": "Diagonal orange-brown planks with vertical posts,
                      receding from the lower right toward the horizon",
     "meaning": "The path along which the figure stands; mentioned in
                Munch's poem",
     "evidence_type": "indirect_evidence"},
    {"object": "Two background figures",
     "construction": "Small figures in cooler dark tones, walking away
                      along the straight path on the left shore",
     "meaning": "Their departure emphasizes the isolation of the
                foreground figure; the 'two friends' of the poem",
     "evidence_type": "direct_evidence"}
  ]
}
\end{verbatim}
\end{footnotesize}

\subsubsection{Documentary Evidence}

\begin{footnotesize}
\begin{verbatim}
{
  "documentary_evidence": {
    "sources": [
      {"source": "Munch's prose poem",
       "date": "1892",
       "significance": "Directly describes the scene: the sky turning
                        blood-red above the blue-black fjord while Munch
                        stood trembling with anxiety and sensed an
                        infinite scream passing through nature; two
                        friends are described walking on along the shore",
       "evidence_type": "direct_evidence"},
      {"source": "Upper-left inscription and infrared/handwriting analysis",
       "date": "c. 1904",
       "significance": "The Norwegian inscription 'Kan kun v\ae re malet
                        af en gal Mand!' ('Can only be painted by a madman!')
                        was revealed by infrared imaging and confirmed by
                        handwriting analysis to be Munch's own",
       "evidence_type": "direct_evidence"}
    ]
  }
}
\end{verbatim}
\end{footnotesize}

\subsubsection{Provenance}

\begin{footnotesize}
\begin{verbatim}
{
  "provenance": {
    "chain_of_ownership": [
      {"period": "1893 - 1910",
       "owner": "Edvard Munch",
       "transaction": "Painted in 1893; first exhibited the same year in
                       Berlin under the title Verzweiflung ('Despair')",
       "evidence_type": "direct_evidence"},
      {"period": "1910 - present",
       "owner": "Norwegian National Gallery (now part of the National
                 Museum, Oslo)",
       "transaction": "Acquired by gift from collector Olaf Schou in 1910;
                       the gallery had shown interest since 1901",
       "evidence_type": "direct_evidence"}
    ],
    "notes": [
      {"note": "Briefly stolen in 1994, along with Madonna, and recovered;
                custody has otherwise been stable, and it is now on
                permanent display in Oslo",
       "evidence_type": "direct_evidence"}
    ]
  }
}
\end{verbatim}
\end{footnotesize}

\subsubsection{Scientific Analysis}

\begin{footnotesize}
\begin{verbatim}
{
  "scientific_analysis": {
    "pigment_analysis": {
      "finding": "XRF studies (2010-2013) detected cadmium yellow,
                  vermilion, ultramarine, viridian, and other
                  19th-century pigments, confirming Munch's early use of
                  the then-new cadmium yellow",
      "evidence_type": "direct_evidence", 
      "source": "Munchmuseet conservation research; Pettersen et al., 2022"
    },
    "infrared_reflectography": {
      "finding": "Revealed the faint 'madman' inscription in the upper
                  left, once invisible to the eye; hyperspectral and
                  multispectral imaging further mapped pigments and any
                  varnish layers",
      "evidence_type": "direct_evidence", 
      "source": "Munchmuseet conservation research; Nasjonalmuseet (Oslo)"
    },
    "fading_study": {
      "finding": "Conservation science (2017-2020) found humidity, not
                  light exposure, is the primary factor causing the
                  originally bright yellow strokes to fade to white,
                  informing preservation through humidity control",
      "evidence_type": "direct_evidence",
      "source": "Munchmuseet conservation research; Pettersen et al., 2022"
    }
  }
}
\end{verbatim}
\end{footnotesize}

\subsubsection{Interpretations}

\begin{footnotesize}
\begin{verbatim}
{
  "interpretations": {
    "frameworks": [
      {"id": "universal_anxiety",
       "claim": "Broadly read as a symbol of modern anxiety, alienation,
                 or existential despair; because the figure has no fixed
                 identity, the suffering reads as universal",
       "evidence_type": "interpretation"},
      {"id": "contemporary_readings",
       "claim": "Early critics such as Przybyszewski (1894) read a
                 sexual struggle ('a battle between brain and sex'),
                 while later analysts favor a universal-angst reading",
       "conflict_flag": true,
       "evidence_type": "interpretation"},
      {"id": "psychoanalytic_reading",
       "claim": "Links between the scream and Munch's own neurosis or
                 genetic predisposition have been proposed but remain
                 speculative",
       "evidence_type": "speculative"}
    ]
  }
}
\end{verbatim}
\end{footnotesize}

\subsubsection{Cultural Impact}

\begin{footnotesize}
\begin{verbatim}
{
  "cultural_impact": {
    "recognition": {
      "claim": "Among the most recognized images in art, widely regarded
                as a universal icon of dread and a watershed of early
                Expressionism",
      "evidence_type": "interpretation"
    }
  }
}
\end{verbatim}
\end{footnotesize}

\subsubsection{Conflicts}

\begin{footnotesize}
\begin{verbatim}
"conflicts": [
    {
      "content": "Figure identity is ambiguous: sources note it is unclear
                  whether the central figure is male or female, clothed or
                  not; Munch intentionally left gender and features
                  indistinct.",
      "evidence_type": "interpretation",
      "sources": [
        "Nasjonalmuseet (Oslo) collection catalogue, NG.M.00939",
        "Published scholarship on The Scream and museum exhibits"
      ]
    },
    {
      "content": "Inscription authorship was long disputed. Most earlier
                  scholars doubted the penciled note was Munch's own, but
                  recent infrared and handwriting analysis attributes it
                  to him.",
      "evidence_type": "direct_evidence",
      "sources": [
        "Nasjonalmuseet (Oslo) collection catalogue, NG.M.00939",
        "Munchmuseet conservation research"
      ]
    },
    {
      "content": "Version dating was historically confused between the painted versions and pastels. It is now accepted that
     the Oslo National Museum painting is 1893 and the Munch
    Museum version is 1910.",
      "evidence_type": "interpretation",
      "sources": [
        "Munchmuseet official publications and articles",
        "Nasjonalmuseet (Oslo) collection catalogue, NG.M.00939"
      ]
    },
    {
      "content": "Interpretations diverge sharply: early critics such as
                  Przybyszewski (1894) read the work in sexual-symbolic
                  terms, while later critics favor a universal-anxiety
                  theme.",
      "evidence_type": "interpretation",
      "sources": [
        "Published scholarship on The Scream and museum exhibits"
      ]
    }
  ],
}
\end{verbatim}
\end{footnotesize}

\section{Process Synthesis Output}

The complete abductive trajectory $\tau = (s_1, \ldots, s_{N})$ produced. Each step is a quadruple of
\textsc{Observation}, \textsc{Rule}, \textsc{Hypothesis}, \textsc{Action}, and
\textsc{EvidenceKeys}, as specified in \Cref{sec:abductive}.

\subsection{Mona Lisa}
\label{app:trajectory-monalisa}
\begin{footnotesize}
\begin{verbatim}
{
"summary": "The generative process of the 'Mona Lisa' proceeds from
support preparation and compositional planning through
underdrawing, background development, figure construction,
costume rendering, sfumato modeling, and final refinements
of the face, hands, lighting, and landscape.",

"steps": [
{
"step": 1,
"observation": "The painting is executed on a thin poplar wood panel,
indicating careful preparation of the support.",
"rule": "For a High Renaissance portrait of this scale, a prepared
poplar panel would provide a stable and suitable surface
for oil painting.",
"hypothesis": "The painter began by preparing a large, thin poplar
panel as the support for the portrait.",
"action": "support_preparation: Prepare a large thin poplar wood
panel suitable for a near life-size half-length portrait.",
"memory_keys": [
"mona_lisa_knowledge_full.materials_and_technique"
]
},

{
  "step": 2,
  "observation": "The composition centers a half-length seated figure,
                  turned slightly to one side while facing the viewer,
                  with head and upper body occupying the central field.",
  "rule": "The compositional arrangement of a frontal seated figure
           suggests a deliberate planning phase to establish the main
           portrait format.",
  "hypothesis": "The painter first established the main portrait
                 arrangement, placing the figure centrally before
                 proceeding to other elements.",
  "action": "planning: Establish the main portrait arrangement: a
             seated woman in a rounded-backed chair, turned slightly
             and placed before a balustrade with landscape beyond.",
  "memory_keys": [
    "mona_lisa_knowledge_full.ematerials_and_technique"
  ]
},

{
  "step": 3,
  "observation": "Infrared reflectography reveals underlying structures
                  such as a rounded-backed chair, balusters, and a
                  balustrade, indicating initial compositional planning.",
  "rule": "Infrared evidence of compositional details beneath the
           surface suggests the use of underdrawing or compositional
           sketches before full paint application.",
  "hypothesis": "The painter likely used a preparatory underdrawing to
                 define the main structural elements of the composition.",
  "action": "underdrawing: Sketch the underlying composition using
             charcoal or similar medium, including the rounded-backed
             chair, balusters, balustrade, and distant landscape
             features.",
  "memory_keys": [
    "mona_lisa_knowledge_full.scientific_analysis"
  ]
},

{
  "step": 4,
  "observation": "The distant landscape features mountain chains, water,
                  a winding road, and a bridge, indicating an atmospheric
                  background.",
  "rule": "A detailed atmospheric background requires careful rendering
           of distant elements to create depth and recession.",
  "hypothesis": "The painter started with the background to establish
                 the spatial context before focusing on the figure.",
  "action": "underpainting: Develop the background setting with distant
             mountains, watercourses, a winding road on the left, and a
             bridge over a river on the right.",
  "memory_keys": [
    "mona_lisa_knowledge_full.formal_analysis"
  ]
},

{
  "step": 5,
  "observation": "The sitter's main figure, including torso, arms, and
                  face, is clearly defined with soft transitions.",
  "rule": "The main figure's structure must be established before
           detailed rendering of facial features, costume, and
           accessories.",
  "hypothesis": "The painter constructed the main figure next, focusing
                 on torso, arms, and basic facial structure.",
  "action": "main_layers: Construct the sitter's main figure, including
             torso, arms, and basic facial structure, using muted earth
             tones and initial tonal gradations.",
  "memory_keys": [
    "mona_lisa_knowledge_full.formal_analysis"
  ]
},

{
  "step": 6,
  "observation": "The sitter's costume includes a dark dress with visible
                  sleeves and a chemise at the shoulder opening, as
                  revealed by infrared reflectography.",
  "rule": "Detailed costume elements require careful rendering after
           the main figure structure is established.",
  "hypothesis": "The painter added costume details following the main
                 figure construction.",
  "action": "main_layers: Render costume details including a dark dress,
             likely yellow detachable sleeves, and a white chemise at
             the shoulder opening.",
  "memory_keys": [
    "mona_lisa_knowledge_full.formal_analysis"
  ]
},

{
  "step": 7,
  "observation": "The transparent veil is a prominent feature, covering
                  the head and shoulders and descending over the torso.",
  "rule": "The veil's transparency and placement suggest it was
           integrated with the figure's hair and clothing, requiring
           careful layering.",
  "hypothesis": "The painter added the veil after the basic figure and
                 costume were established.",
  "action": "main_layers: Paint a transparent veil over the head and
             shoulders, descending over the torso and arms, using
             subtle gradations to achieve transparency.",
  "memory_keys": [
    "mona_lisa_knowledge_full.iconographic_analysis"
  ]
},

{
  "step": 8,
  "observation": "The hands are folded across the lower foreground,
                  forming a stable base for the figure.",
  "rule": "Carefully rendered hands require focused attention after
           the main figure and costume are in place.",
  "hypothesis": "The painter added the folded hands next, ensuring they
                 complement the overall composition.",
  "action": "main_layers: Place the hands across the lower foreground,
             forming the stable base of the figure, with careful
             attention to soft modeling and tonal transitions.",
  "memory_keys": [
    "mona_lisa_knowledge_full.iconographic_analysis"
  ]
},

{
  "step": 9,
  "observation": "The mouth is softly modeled with blurred transitions
                  at the corners, avoiding a hard outline.",
  "rule": "Soft sfumato effects around the mouth require careful
           blending and multiple glazing layers.",
  "hypothesis": "The painter refined the mouth area with multiple layers
                 to achieve the characteristic sfumato effect.",
  "action": "upper_layers: Model the mouth with soft tonal gradations,
             using blurred transitions at the corners to create the
             subtle smile.",
  "memory_keys": [
    "mona_lisa_knowledge_full.formal_analysis"
  ]
},

{
  "step": 10,
  "observation": "The eyes face outward toward the viewer, framed by soft
                  shadows and hair, with subtle highlights and brown
                  shadow transitions.",
  "rule": "Soft sfumato effects around the eyes require careful
           blending and attention to tonal transitions.",
  "hypothesis": "The painter refined the eyes with multiple layers to
                 achieve the characteristic sfumato effect and direct
                 gaze.",
  "action": "upper_layers: Model the eyes with soft tonal gradations,
             adding subtle highlights and shadow transitions to create
             the direct gaze and sfumato effect.",
  "memory_keys": [
    "mona_lisa_knowledge_full.formal_analysis"
  ]
},

{
  "step": 11,
  "observation": "The overall lighting is soft, with gentle illumination
                  on the face, chest, and hands, avoiding strong cast
                  shadows.",
  "rule": "Soft atmospheric lighting requires careful modulation of
           light and shadow across the entire painting.",
  "hypothesis": "The painter refined the lighting effects across the
                 figure and background to achieve a harmonious and
                 tranquil atmosphere.",
  "action": "upper_layers: Refine the overall lighting, ensuring soft
             illumination on the face, chest, and hands, with gentle
             shading and atmospheric recession in the background.",
  "memory_keys": [
    "mona_lisa_knowledge_full.formal_analysis"
  ]
},

{
  "step": 12,
  "observation": "The distant landscape features a winding road and
                  bridge, contributing to the sense of depth and
                  recession.",
  "rule": "Fine details in the distant landscape require careful
           attention to maintain atmospheric perspective and tonal
           gradations.",
  "hypothesis": "The painter refined the distant landscape details to
                 enhance depth and harmony with the overall composition.",
  "action": "upper_layers: Refine the distant landscape details,
             including the winding road and bridge, using soft tonal
             transitions and atmospheric perspective.",
  "memory_keys": [
    "mona_lisa_knowledge_full.formal_analysis"
  ]
},

{
  "step": 13,
  "observation": "The face is the focal point, with careful modeling
                  around the eyes, nose, and cheeks, achieving a lifelike
                  quality.",
  "rule": "The face requires multiple layers of refinement to achieve
           the subtle modeling and psychological ambiguity seen in the
           portrait.",
  "hypothesis": "The painter spent considerable time refining the face,
                 particularly around the eyes, nose, and cheeks.",
  "action": "final_accents: Refine the face with multiple layers,
             focusing on subtle modeling around the eyes, nose, and
             cheeks, to achieve lifelike qualities and psychological
             ambiguity.",
  "memory_keys": [
    "mona_lisa_knowledge_full.formal_analysis"
  ]
},

{
  "step": 14,
  "observation": "The hands are carefully modeled with warm flesh tones
                  and soft transitions, anchoring the lower composition.",
  "rule": "Final refinements to the hands ensure they complement the
           overall composition and contribute to the sense of composure
           and stillness.",
  "hypothesis": "The painter refined the hands with additional layers to
                 achieve the desired soft modeling and tonal harmony.",
  "action": "final_accents: Refine the hands with additional layers,
             ensuring warm flesh tones and soft transitions that
             complement the overall composition.",
  "memory_keys": [
    "mona_lisa_knowledge_full.formal_analysis"
  ]
},

{
  "step": 15,
  "observation": "The distant landscape maintains a harmonious balance
                  with the figure, achieved through soft atmospheric
                  effects.",
  "rule": "Final adjustments to the background ensure it complements
           the figure and contributes to the overall mood of the
           painting.",
  "hypothesis": "The painter made final adjustments to the distant
                 landscape to ensure it harmonizes with the figure and
                 maintains atmospheric depth.",
  "action": "synthesis: Make final adjustments to the distant landscape,
             ensuring it complements the figure and maintains
             atmospheric depth and tranquility.",
  "memory_keys": [
    "mona_lisa_knowledge_full.formal_analysis"
  ]
},

{
  "step": 16,
  "observation": "The overall tone and color balance of the painting is
                  muted and warm-brown, indicative of aged varnish
                  layers.",
  "rule": "The current color balance is altered by thick, irregular,
           oxidized varnish layers applied after Leonardo's death.",
  "hypothesis": "The painter's original intention may have included
                 cooler tones and greater contrast, but the current
                 appearance reflects later varnish changes.",
  "action": "post_completion: Apply varnish layers consistent with
             historical evidence, understanding that these alter the
             original palette but are part of the painting's current
             appearance.",
  "memory_keys": [
    "mona_lisa_knowledge_full.scientific_analysis"
  ]
}

]
}
\end{verbatim}
\end{footnotesize}

\subsection{Cafe Terrace at Night}
\label{app: cafe_process}
\begin{footnotesize}
\begin{verbatim}
{
"summary": "The generative process of 'Cafe Terrace at Night' proceeds from
compositional planning and underdrawing through underpainting,
main-layer construction of the terrace, figures, and sky,
impasto texturing, and final refinements of color balance
and overall synthesis.",
"steps": [
{
"step": 1,
"observation": "The artwork features a portrait orientation with strong
                one-point perspective, dominated by a glowing yellow
                cafe terrace on the left and a deep blue starry sky
                above.",
"rule": "For a faithful reconstruction, the initial underdrawing should
         establish the spatial relationships and proportions
         described in the formal analysis.",
"hypothesis": "The painter began by sketching the basic composition,
               including the cafe terrace, cobblestone street,
               buildings, and tree, ensuring accurate placement
               and perspective.",
"action": "planning: Create a detailed underdrawing with a light
           pencil, outlining the cafe terrace extending from the
           bottom left, the cobblestone street receding diagonally
           toward a dark vanishing point, and the starry sky above.
           Include the awning, standing waiter, and key elements
           like buildings and the tree.",
"memory_keys": [
  "cafe_terrace_knowledge_full.formal_analysis.composition",
  "cafe_terrace_knowledge_full.iconographic_analysis"
]
},
{
  "step": 2,
  "observation": "The color palette emphasizes warm yellows and oranges
                  for the terrace and cooler blues and violets for the
                  night sky, with no black used.",
  "rule": "The underpainting should set the overall tone and color
           harmony, building the foundation for subsequent layers.",
  "hypothesis": "The painter applied a thin, translucent layer of warm
                 yellows and oranges for the cafe terrace area and
                 cooler blues and violets for the sky and buildings.",
  "action": "underpainting: Paint a thin, translucent layer of warm
             yellow and orange hues for the terrace and awning,
             blending into cooler blue and violet tones for the sky
             and buildings. Leave the cobblestone street in a muted
             pinkish-purple shade, catching reflected light.",
  "memory_keys": [
    "cafe_terrace_knowledge_full.color_and_light"
  ]
},
{
  "step": 3,
  "observation": "The cobblestone street has agitated short strokes,
                  and the terrace features smoother passages,
                  indicating varied brushwork techniques.",
  "rule": "Different textures and brushstrokes should be applied to
           capture the distinct surfaces, as described in the
           materials and techniques.",
  "hypothesis": "The painter used agitated short strokes for the
                 cobblestone street and smoother, broader strokes for
                 the terrace and awning.",
  "action": "main_layers: Layer agitated short strokes of violet-pink
             and blue-violet hues for the cobblestone street, applying
             impasto texture. For the terrace, apply smoother, broader
             strokes of warm yellow and orange, gradually blending
             into shadows.",
  "memory_keys": [
    "cafe_terrace_knowledge_full.materials_and_technique"
  ]
},
{
  "step": 4,
  "observation": "The standing waiter in white is positioned near the
                  lit doorway, acting as a vertical pivot in the
                  composition.",
  "rule": "Key figures should be placed accurately to maintain
           compositional integrity.",
  "hypothesis": "The painter positioned the standing figure in white
                 near the lit doorway, ensuring it stands out against
                 the warm yellow backdrop of the terrace.",
  "action": "main_layers: Paint the standing waiter in bright white,
             using smooth, clean strokes to create a stark contrast
             against the warm yellow tones of the terrace. Ensure the
             figure is centrally aligned near the lit doorway.",
  "memory_keys": [
    "cafe_terrace_knowledge_full.iconographic_analysis"
  ]
},
{
  "step": 5,
  "observation": "The starry night sky is dotted with swirled yellow
                  stars, creating a dynamic contrast against the deep
                  blue and violet tones.",
  "rule": "Stars should be depicted with swirled white touches to
           capture their luminous quality.",
  "hypothesis": "The painter added swirled white touches for the stars
                 across the upper third of the canvas, particularly
                 concentrated in the upper-right area.",
  "action": "main_layers: Using a fine brush, add swirled white touches
             for stars across the deep blue and violet sky,
             concentrating them in the upper-right corner. Ensure they
             catch light and stand out against the cool background.",
  "memory_keys": [
    "cafe_terrace_knowledge_full.iconographic_analysis"
  ]
},
{
  "step": 6,
  "observation": "The tree on the right side is a dark green-black mass
                  with bristling boughs, partially obscuring the
                  buildings behind it.",
  "rule": "The tree should be painted with dark green-black tones and
           bristling strokes to convey its texture and form.",
  "hypothesis": "The painter rendered the tree with dark green-black
                 tones, using bristling strokes to depict its foliage
                 and branches, positioned to partially obscure the
                 buildings behind it.",
  "action": "main_layers: Apply dark green-black paint to the tree
             area, using bristling, textured strokes to simulate the
             boughs and leaves. Ensure it extends into the upper-right
             corner, overlapping parts of the sky and buildings.",
  "memory_keys": [
    "cafe_terrace_knowledge_full.iconographic_analysis"
  ]
},
{
  "step": 7,
  "observation": "Tables, chairs, and patrons are arranged on the
                  terrace, contributing to the warm social atmosphere.",
  "rule": "Figures and objects should be loosely blocked to maintain
           the impressionistic style, focusing on their shapes and
           positions rather than precise details.",
  "hypothesis": "The painter added rough, block-like forms for tables,
                 chairs, and patrons, using warm lit tones for the
                 seated figures and darker shadows for those further
                 back.",
  "action": "main_layers: Using warm yellow and orange hues, paint
             rough forms for tables and chairs on the terrace. Add
             seated figures with simple, block-like shapes,
             incorporating lighter highlights for those closer to the
             viewer.",
  "memory_keys": [
    "cafe_terrace_knowledge_full.iconographic_analysis"
  ]
},
{
  "step": 8,
  "observation": "Pedestrians strolling along the cobblestone street are
                  small, shadowy figures, adding scale and movement to
                  the scene.",
  "rule": "Pedestrians should be rendered simply, emphasizing their
           role as scale markers rather than detailed individuals.",
  "hypothesis": "The painter added small, dark silhouettes for
                 pedestrians along the cobblestone street, using
                 quick, gestural strokes to suggest movement and
                 depth.",
  "action": "main_layers: Along the diagonal cobblestone street, paint
             small, dark figures using quick, gestural strokes. Ensure
             they fade into shadow as they recede into the distance.",
  "memory_keys": [
    "cafe_terrace_knowledge_full.iconographic_analysis"
  ]
},
{
  "step": 9,
  "observation": "The impasto technique is evident throughout the
                  painting, particularly in the sky and street areas,
                  enhancing the textured surface.",
  "rule": "Impasto texture should be applied generously to mimic the
           original's material richness and visual interest.",
  "hypothesis": "The painter added thick, textured impasto layers to
                 the sky and street, especially in the cobblestones
                 and starry areas, to enhance depth and luminosity.",
  "action": "upper_layers: Build up thick impasto layers in the sky
             and cobblestone street areas, using a palette knife to
             apply the paint. Focus on the cobblestones with agitated
             short strokes and the stars with swirled touches.",
  "memory_keys": [
    "cafe_terrace_knowledge_full.materials_and_technique"
  ]
},
{
  "step": 10,
  "observation": "The overall scene maintains a warm-cool color
                  contrast, with the yellow terrace and awning
                  providing a focal point against the cool night sky.",
  "rule": "Final adjustments should refine the color balance and
           ensure the warm-cool contrast is maintained throughout the
           painting.",
  "hypothesis": "The painter reinforced the warm-cool contrast by
                 adjusting the saturation and brightness of the yellow
                 terrace and the cool tones of the sky and buildings.",
  "action": "final_accents: Enhance the warm yellow and orange hues of
             the terrace and awning, making them slightly brighter and
             more saturated. Deepen the cool blue and violet tones of
             the sky and buildings, ensuring they complement the warm
             elements without overpowering them.",
  "memory_keys": [
    "cafe_terrace_knowledge_full.color_and_light"
  ]
},
{
  "step": 11,
  "observation": "The final image should convey the essence of Van
                  Gogh's style, including his signature brushwork and
                  vibrant color palette.",
  "rule": "The synthesis stage ensures all elements work together to
           create a cohesive and dynamic composition.",
  "hypothesis": "The painter reviewed the entire composition, refining
                 details and ensuring the overall balance of warm and
                 cool tones, textures, and spatial relationships.",
  "action": "synthesis: Step back and review the painting, making
             final adjustments to ensure the warm yellow terrace and
             awning dominate the left and center, contrasting sharply
             with the cool blue and violet night sky. Refine the
             details of the cobblestones, buildings, and tree to
             match the original's intricate level of detail.",
  "memory_keys": [
    "cafe_terrace_knowledge_full.formal_analysis",
    "cafe_terrace_knowledge_full.iconographic_analysis"
  ]
}
]
}
\end{verbatim}
\end{footnotesize}
\section{Prompt Templates}
\label{app:prompt-templates}
\subsection{Abduction Agent Prompt}
\begin{promptbox}{Abduction Agent}
\begin{footnotesize}
\begin{verbatim}
ABDUCTION AGENT

ROLE
You are an art historian specializing in process reconstruction. Given a
reference artwork and its structured evidence repository M_evidence, infer the
ordered sequence of production steps that most plausibly explains how the work
was made.

STEP 0 - LOOK FIRST
Before reading the evidence, observe the reference image directly. Note the
visible elements, their layout, palette, and surface character. Treat these
observations only as features to be explained; the explanation must come from
M_evidence, not from appearance alone.

{POLICY}

READING THE EVIDENCE (reliability priority)
Each claim in M_evidence carries exactly one reliability tag. When grounding a
step, prefer the strongest tier that applies, in this order:
  1. direct_evidence  : primary sources, museum records, artist correspondence,
                        conservation and laboratory reports. Use to fix support,
                        materials, documented structure, and order.
  2. indirect_evidence: claims inferred from imaging or visual analysis. Use for
                        structure and technique not stated in a primary record.
  3. interpretation   : scholarly readings of meaning, style, or intent. Use to
                        motivate stylistic choices, not to assert material fact.
  4. speculative      : weakly supported or contested claims. Use only when no
                        stronger evidence bears on the step.
Do not justify a material or structural action with an interpretation or
speculative claim when a direct or indirect claim is available. Any step
grounded only in interpretation or speculative evidence must be flagged
low-confidence.

READING CONFLICTS
When the key you would cite carries a conflict_flag, or the topic appears in the
conflicts field, do NOT merge or average the competing positions. Select the
position supported by the higher-reliability tag or by the stated scholarly
consensus, ground the action in that position, and record the contested status.
Example: where the darkened costume is read as either mourning dress or aged
varnish, follow the conservation reading and treat the darkening as a later
varnish effect, not an original color decision. For conflicts marked
unresolved, keep the action underspecified rather than committing to a
contested detail.


OUTPUT
Return a JSON list of steps. Each step is a quadruple
{observation, rule, hypothesis, action} with memory_keys listing the cited
evidence keys and the reliability tag carried by each.
\end{tcolorbox}
\captionof{figure}{Abduction agent prompt. The agent observes the reference
image first, then grounds each production step in $M_{\mathrm{evidence}}$ under
a strict reliability ordering (\texttt{direct\_evidence} >
\texttt{indirect\_evidence} > \texttt{interpretation} > \texttt{speculative}),
flagging steps that rest only on the weaker tiers. Conflicted keys are resolved
by the higher-reliability position or the stated consensus rather than averaged,
and unresolved conflicts are left underspecified.}


\end{verbatim}
\end{footnotesize}
\end{promptbox}

\subsection{Deep Research Prompt}

\begin{promptbox}{Deep Research}
\begin{footnotesize}
\begin{verbatim}
OBJECTIVE
Gather comprehensive, source-grounded evidence about one artwork across all
dimensions below, with particular attention to HOW the work was physically
made: the order, stages, tools, material build-up, decisions, and revisions,
as far as the evidence supports. You retrieve and report only; a later stage
will structure the result. Every non-trivial claim must carry an evidence
type and at least one source.

ARTWORK: "{artwork_title}"   ARTIST: "{artist_name}"

STEP 1 - ARTWORK IDENTIFICATION
Resolve the exact object before researching. Confirm canonical title,
alternate titles, artist, year/date, catalogue raisonne numbers (including
F/JH where applicable), holding institution, accession number, and medium /
support / dimensions. If the title is ambiguous or matches multiple works,
list the candidates, select the most likely object, and state that the rest
of the research concerns that object.

STEP 2 - COVERAGE
Search broadly and find evidence for every dimension:
  A. Basic metadata
  B. Historical context (life circumstances, working conditions, lighting /
     time-of-day, health, finances, studio/outdoor context, socio-political
     and art-historical context)
  C. Material and technical analysis (support and preparation; pigments,
     binders, color change; layer order and build-up; wet-in-wet vs
     successive applications; sittings; tools and marks; corrections and
     reworkings; finishing actions; conservation findings)
  D. Formal analysis (form, composition, color, light, perspective, space)
  E. Iconographic analysis (each important object and motif)
  F. Documentary evidence (letters, journals, recorded statements, museum
     and archival records; cite primary letter numbers where they exist)
  G. Provenance and ownership history
  H. Scientific and analytical insights (IRR, X-radiography, raking light,
     MA-XRF, OCT, cross-sections, pigment analysis, conservation science,
     astronomical/physics studies); prioritize findings that reveal
     subsurface layers, underdrawing, order of application, and alterations
  I. Interpretations (scholarly readings, competing viewpoints, criticism)
  J. Cultural impact and reception
  K. Execution and construction (discrete making steps from support
     preparation to final marks, with dependencies where known)
  L. Related and preparatory works (sketches, studies, letters of intent,
     replicas/later versions, position within a series)

SOURCE ATTRIBUTION
Attach at least one source to every non-trivial claim. Prefer holding-museum
records, artist letters, catalogues raisonnes, peer-reviewed conservation /
science journals, and reputable scholarly sources. Avoid low-credibility
sources. Capture, per item, the exact claim or passage, the specific source
(institution, author, document, date), and a URL or citation when available.

CONFLICTS
Where sources disagree on date, title, dimensions, provenance,
interpretation, layer order, or whether a feature is original / altered /
restored, retain ALL major viewpoints with their sources. Never pick a
winner here.

DISCIPLINE
Do not invent facts, fill gaps, or fabricate a complete making sequence.
Leave ordering open where evidence is silent. Report undocumented topics as
undocumented.

OUTPUT
Return a flat list of retrieved items, each as
  {claim, dimension, evidence_type, source, citation_or_url}.
Group nothing and resolve nothing; raw retrieved items only.
\end{verbatim}
\end{footnotesize}
\end{promptbox}

\subsection{Schema-Filling Prompt}

\begin{promptbox}{Schema Filling}
\begin{footnotesize}
\begin{verbatim}
OBJECTIVE
Organize the raw evidence retrieved for one artwork into the fixed schema
below as STRICT JSON. Place each claim in the correct field, tag its
reliability, attach its source, and preserve disagreements. Add no facts of
your own. Every non-trivial claim must have an evidence type and at least
one source.

ARTWORK: "{artwork_title}"   ARTIST: "{artist_name}"
RETRIEVED EVIDENCE (JSON): {retrieved_items}

EVIDENCE TAGGING (exactly one per entry; do not mix; split entries if needed)
  direct_evidence  : primary sources, artist letters, museum records,
                     lab / conservation reports
  indirect_evidence: analysis from observation, imaging, or technical data
  interpretation   : expert or scholarly opinion
  speculative      : weakly supported or uncertain claims

CONFLICT HANDLING
If sources disagree, keep the consistent claims in their field and record the
competing viewpoints explicitly in the "conflicts" section, each with its
source. Do not merge or average competing claims.

ORDERING & UNCERTAINTY DISCIPLINE
Assert that one action/layer/mark precedes another only with support from
stratigraphy, imaging, a conservation report, the artist's own account, or
cited scholarship. Otherwise set ordering_note to null/"unknown" and
confidence_in_ordering to "low" or "unknown". Never invent stages to make
the process look complete; partial ordering is preferred.

ENTRY SCHEMAS
  A. STANDARD (historical_context, documentary_evidence, provenance,
     interpretations, cultural_impact, conflicts,
     related_and_preparatory_works):
     {"content","evidence_type","sources":[]}
  B. LAYER-AWARE (materials_and_technique, scientific_analysis):
     {"content","evidence_type",
      "stage":"support_preparation|underlayer|main_layers|upper_layers|
               final_accents|preparatory|post_completion|unknown",
      "ordering_note":"what this was applied over/under or followed;
                       null if undetermined","sources":[]}
  C. FORMAL (each key of formal_analysis):
     {"content","evidence_type","sources":[]}
  D. ICONOGRAPHIC (iconographic_analysis only):
     {"object","object_identity","visual_construction","color_and_light",
      "symbolic_meaning","functional_role","evidence_type","sources":[]}


COMPLETENESS CHECK (before returning)
  - no major category missing;
    both present
  - no redundancy across sections
  - every making step traceable to at least one source
  - every entry has evidence_type and sources
  - top-level "sources" lists every source used
  - output is valid JSON only

OUTPUT FORMAT (strict JSON only)
{"artwork":{"title":"","alternate_titles":[],"artist":"","year":"",
"metadata":{"medium":"","support":"","dimensions_cm":"","dimensions_in":"",
"location":"","holding_institution":"","accession_number":"",
"credit_line":"","catalogue_numbers":[]}},
"historical_context":[],"materials_and_technique":[],
"formal_analysis":{"form":{},"composition":{},"color":{},"light":{},
"perspective":{},"space":{}},
"iconographic_analysis":[],"documentary_evidence":[],"provenance":[],
"scientific_analysis":[],"execution_and_construction":[],
"related_and_preparatory_works":[],"interpretations":[],
"cultural_impact":[],"conflicts":[],"sources":[]}
\end{verbatim}
\end{footnotesize}
\end{promptbox}
\subsection{DAG Construction Prompt}
\label{app:dag-construction-prompt}


\begin{promptbox}{Composition Agent}
Below are the ordered steps of a painting process. For EACH step, extract the
single main visual ELEMENT it puts on the canvas and ALL of its attributes that
are stated or clearly implied. Also give the edges: which earlier element each
one builds on, the relation, and the position of this element relative to that
one.

STEPS (JSON): \{...\}

\textbf{Return format.} Return JSON with two keys.
\begin{itemize}
  \item \texttt{nodes}: a LIST, one per step, each
  \{\texttt{id}: int (matching the step id), \texttt{object}: short noun for
  the element, \texttt{element}: a concise finished-appearance phrase (no verbs
  like `apply'), \texttt{attributes}: an OBJECT mapping attribute TYPE to a
  list of short values. Use as many types as apply: color, size, position,
  brushstroke, texture, lighting, shape, material, mood. Preserve any explicit
  coordinates (e.g. `\textasciitilde60\% width'). Extract EVERYTHING present;
  do not invent.\}
  \item \texttt{edges}: a LIST of \{\texttt{from}: int,
  \texttt{to}: int, \texttt{relation}: one of
  over/refining/accenting/reflecting,
  \texttt{relative\_position}: a short phrase like
  `to the lower-left of'\}. Base edges on the build dependencies
  (a step's \texttt{follows}).
\end{itemize}

No prose, JSON only.
\end{promptbox}
\subsection{DAG to prompt construction}
\label{app:prompt-construction-prompt}

\begin{promptbox}{Visual Prompt  Agent}
You are given a process graph for a painting, node-wise. Each node is one visual
element with its full attribute set (colour, size, brushwork, position, and any
others). One node is the anchor (\texttt{is\_anchor=true}); higher importance =
more visually dominant. Each edge gives the compositional relation and the
position of the target element relative to the source.

GRAPH (JSON): \{...\}

From this graph, write the FINISHED-painting description for a text-to-image
model. Requirements:

\begin{itemize}
  \item Use the attributes given on each node (colours, sizes, brushwork); do not drop them.
  \item Make the anchor the dominant focal element and lead with it; scale prominence by importance.
  \item Place every other element using its edge relation and relative position.
  \item Express POSITION qualitatively (upper-left, lower-center, near the horizon); do NOT use percentage coordinates for position. Percentages may ONLY describe SIZE.
  \item Do NOT begin with a palette or label line; weave any colour/atmosphere note into the description.
  \item Describe only the final APPEARANCE, never the steps or verbs like `apply'.
  \item Do NOT name any artist or artwork title; use no proper names.
  \item Be detailed; 4--8 sentences is fine.
\end{itemize}

\textbf{Return format.}
Return JSON with one key \texttt{final\_image\_prompt}: a detailed appearance description that uses every node's attributes and the edge positions, no proper names, no step language.
\end{promptbox}
\subsection{Self-Reflection Prompt}
\label{app:self-reflection-prompt}

\begin{promptbox}{Self-Reflection}
Image 1 is the ORIGINAL artwork (target). Image 2 is what your process produced.

You are given the three similarity rewards with the weakest axis named, the
documented elements to check, the ordered process actions that produced Image 2,
and the knowledge base.

Do TWO things:
(1) compare the two images element by element;
(2) judge how well the process used the knowledge base, then propose corrections
in two categories, grounded in the knowledge base.

\textbf{Return format.}

Return JSON with these fields:

\begin{itemize}
  \item \texttt{differences}: a LIST, one entry per differing element, each
  \{\texttt{element}, \texttt{attribute}
  (presence/color/size/position/shape/texture),
  \texttt{reference},
  \texttt{generated},
  \texttt{severity} in [0,1]\}.

  \item \texttt{process\_critique}: 1--3 sentences on how the process mis-used the knowledge base.

  \item \texttt{axis}: the dominant (weakest) mismatch axis, one of
  material / procedural / intentional.

  \item \texttt{violated\_principle}: one sentence on the biggest divergence.

  \item \texttt{severity}: float in [0,1], the maximum element severity.

  \item \texttt{observation}: a short generic summary
  (no title/subject/specific colours).

  \item \texttt{generic\_constraints}: a LIST of generic art-creation-process
  rules that apply to any artwork.

  \item \texttt{artwork\_constraints}: a LIST of corrections specific to this
  work, grounded in the knowledge base.

  \item \texttt{suggested\_role\_refinement}: optional generic
  medium/technique qualifier; empty if not needed.
\end{itemize}

If the two images are essentially identical, return severity 0.0 and empty lists.
\end{promptbox}
\begin{center}
\begin{minipage}{\columnwidth}

\begin{methodpromptbox}{CoTColor}{\texttt{CoT prompt}}
\tiny
\textbf{Goal:} infer a step-by-step production process from the reference image.

\vspace{0.2em}

\textbf{Prompt structure:}
\begin{itemize}[leftmargin=1.1em,itemsep=0.05em,topsep=0pt,parsep=0pt]
    \item Think step by step in a brief \texttt{REASONING} block.
    \item Commit to one ordered \texttt{PROCESS}.
    \item Cover support and ground, composition blocking, spatial layout, palette, lighting, texture, subject build order, and final atmosphere.
    \item Use only visible evidence from the image.
\end{itemize}

\vspace{0.2em}

\textbf{Shared rules:} do not mention the artwork name, artist name, or ``in the style of''; avoid historical or biographical details; make every step concrete enough for image generation.
\end{methodpromptbox}

\captionof{figure}{
CoT baseline prompt. The model first reasons about the visible image and then produces a single ordered production process.
}
\label{fig:prompt_cot}

\end{minipage}
\end{center}
\begin{center}
\begin{minipage}{\columnwidth}

\begin{methodpromptbox}{CoTSCColor}{\texttt{CoT-SC prompt}}
\tiny
\textbf{Goal:} obtain a consensus production process from multiple independent samples.

\vspace{0.2em}

\textbf{Prompt structure:}
\begin{itemize}[leftmargin=1.1em,itemsep=0.05em,topsep=0pt,parsep=0pt]
    \item Sample $k$ independent production-process reconstructions.
    \item Merge candidates by self-consistency.
    \item Keep steps that recur across most candidates.
    \item Keep unique steps only if clearly visible in the image.
    \item Drop weak, contradictory, or speculative steps.
    \item Return one consensus process in correct build order.
\end{itemize}

\vspace{0.2em}

\textbf{Shared rules:} do not mention the artwork name, artist name, or ``in the style of''; use only visible image evidence; describe the process in production order.
\end{methodpromptbox}

\captionof{figure}{
CoT-SC baseline prompt. Multiple independent process reconstructions are sampled and merged into a self-consistent consensus process.
}
\label{fig:prompt_cot_sc}

\end{minipage}
\end{center}
\begin{center}
\begin{minipage}{\columnwidth}

\begin{methodpromptbox}{ToTColor}{\texttt{ToT prompt}}
\tiny
\textbf{Goal:} compare alternative production plans before expanding the selected plan.

\vspace{0.2em}

\textbf{Prompt structure:}
\begin{itemize}[leftmargin=1.1em,itemsep=0.05em,topsep=0pt,parsep=0pt]
    \item Propose three candidate plans.
    \item Plan A: organized around geometric layout and composition.
    \item Plan B: organized around main subjects and their relationships.
    \item Plan C: organized around palette, lighting, texture, and surface.
    \item Give each plan an outline, strength, and weakness.
    \item Score each plan for fidelity and build order.
    \item Expand only the selected plan into the final production process.
\end{itemize}

\vspace{0.2em}

\textbf{Shared rules:} avoid names, style phrases, history, and biography; use only visible evidence; make steps concrete enough for image generation.
\end{methodpromptbox}

\captionof{figure}{
ToT baseline prompt. The model proposes multiple candidate production plans, evaluates them, and expands the selected plan.
}
\label{fig:prompt_tot}

\end{minipage}
\end{center}
\begin{center}
\begin{minipage}{\columnwidth}

\begin{methodpromptbox}{SelfRefineColor}{\texttt{Self-Refine prompt}}
\tiny
\textbf{Goal:} improve an initial production process through critique and revision.

\vspace{0.2em}

\textbf{Prompt structure:}
\begin{itemize}[leftmargin=1.1em,itemsep=0.05em,topsep=0pt,parsep=0pt]
    \item Draft an initial step-by-step production process.
    \item Critique the draft for missing visible elements.
    \item Check whether the build order is correct.
    \item Identify vague palette, lighting, or texture descriptions.
    \item Remove forbidden names, style phrases, and non-visual content.
    \item Revise the process to fix every critique point.
\end{itemize}

\vspace{0.2em}

\textbf{Shared rules:} the refined process must remain name-free, visible-evidence-grounded, ordered from blank surface to finishing touches, and concrete enough for image generation.
\end{methodpromptbox}

\captionof{figure}{
Self-Refine baseline prompt. The model drafts a production process, critiques it, and revises the process before image generation.
}
\label{fig:prompt_self_refine}

\end{minipage}
\end{center}
\definecolor{MartBack}{HTML}{FFF7ED}   
\definecolor{MartFrame}{HTML}{9A3412}  
\definecolor{MartTitle}{HTML}{FED7AA}  
\definecolor{SunriseFrame}{HTML}{F97316}
\definecolor{SunriseBack}{HTML}{FFF7ED}

\definecolor{GuernicaFrame}{HTML}{525252}
\definecolor{GuernicaBack}{HTML}{F5F5F5}

\definecolor{MonaFrame}{HTML}{059669}
\definecolor{MonaBack}{HTML}{ECFDF5}

\definecolor{CafeFrame}{HTML}{7C3AED}
\definecolor{CafeBack}{HTML}{F5F3FF}

\definecolor{ScreamFrame}{HTML}{DC2626}
\definecolor{ScreamBack}{HTML}{FEF2F2}

\tcbset{
  martmain/.style={
    enhanced,
    breakable,
    width=\textwidth,
    colback=MartBack,
    colframe=MartFrame,
    boxrule=0.65pt,
    arc=1pt,
    left=5pt,
    right=5pt,
    top=5pt,
    bottom=5pt,
    before skip=0pt,
    after skip=0pt,
    boxed title style={
      colback=MartTitle,
      colframe=MartFrame,
      boxrule=0.45pt,
      arc=1pt
    },
    title={\textbf{Artwork-specific memory $\mathcal{M}_{\mathrm{art}}$}},
    fonttitle=\small\bfseries,
    coltitle=black
  },
  artworkbox/.style={
    enhanced,
    breakable,
    boxrule=0.55pt,
    arc=1pt,
    left=5pt,
    right=5pt,
    top=4pt,
    bottom=4pt,
    before skip=4pt,
    after skip=4pt,
    fonttitle=\footnotesize\bfseries,
    coltitle=black
  }
}

\begin{tcolorbox}[martmain]

\small
\textbf{Key:} artwork name \hfill
\textbf{Value:} learned artwork-specific corrective constraints

\vspace{0.4em}

\begin{tcolorbox}[
  artworkbox,
  colback=SunriseBack,
  colframe=SunriseFrame,
  title={\texttt{"Impression, Sunrise"}}
]
\begin{enumerate}[leftmargin=1.2em,itemsep=1pt,topsep=1pt]
    \item Reduce the sun to a small, saturated circular disk positioned slightly right of center in the upper-middle section.
    \item Modify the sun reflection to consist of a vertical chain of short broken strokes in orange, red, and yellow descending below the sun.
    \item Shrink both the foreground and secondary boats to compact, dark blocks: place the foreground boat in the lower-center with two simplified figures, and the secondary boat as a low, horizontal form left of center in the mid-lower water.
    \item Adjust the harbor water to feature loose horizontal strokes in cool turquoise, grey-green, and blue, broken by darker boat forms and warm reflection marks.
    \item Abbreviate the masts and cranes to thin vertical and diagonal teal marks across the left, center, and right background, embedded in the mist.
    \item Diminish the size and prominence of the smoke and vapor plumes, keeping them as soft blue-green plumes rising on the left side with muted edges, merging with the surrounding sky haze.
    \item Simplify the industrial ships to large dark horizontal and diagonal masses on the right mid-ground, partly merged with vertical rigging and haze.
    \item Lighten the sky and clouds to a grey-blue haze with warm orange-pink clouds, especially across the upper band.
    \item Position the signature as a dark cursive inscription along the lower-left edge of the canvas.
\end{enumerate}
\end{tcolorbox}

\begin{tcolorbox}[
  artworkbox,
  colback=GuernicaBack,
  colframe=GuernicaFrame,
  title={\texttt{"Guernica"}}
]
\begin{enumerate}[leftmargin=1.2em,itemsep=1pt,topsep=1pt]
    \item Place the bull in the upper-left corner, serving as a structural anchor and counterweight to the grief below, with a frontal face, massive neck, and planted hooves, above the woman holding the limp child.
    \item Position the mother with the dead child directly beneath the bull, with her head thrown back and mouth open as a sharp wedge, carrying the emotional weight of the left cluster.
    \item Centrally locate the agitated horse, pierced with internal patterns suggesting anatomical stress and surface inscription, serving as the structural hinge of the composition.
    \item Break the fallen warrior into distinct parts across the lower middle base, including a severed arm, broken sword, open hand, and a small sprouting flower, forming the shattered foundation of the composition.
    \item Place the light bulb in the top center, shaped like an eye, radiating jagged light above the horse, capping the composition and competing with the hand-held lamp.
    \item Position the lamp-bearing woman leaning inward from the right-center, extending a hand-held lamp toward the center.
    \item Split the right third of the composition into rushing and burning movement, with a fleeing woman at the lower right and a burning figure with raised arms at the far edge, maintaining their spatial separation.
\end{enumerate}
\end{tcolorbox}

\begin{tcolorbox}[
  artworkbox,
  colback=MonaBack,
  colframe=MonaFrame,
  title={\texttt{"Mona Lisa"}}
]
\begin{enumerate}[leftmargin=1.2em,itemsep=1pt,topsep=1pt]
    \item Reintroduce a transparent veil over the head and shoulders, descending over the shoulders and arms with subtle gradations, ensuring its placement and style align with infrared reflectography findings.
    \item Render the distant landscape to include distant mountains, watercourses, a winding road on the left, and a bridge over a river on the right, using cooler atmospheric tones and soft transitions to deepen the background and create a sense of distance, while avoiding overly bright and saturated colors and architectural elements like buildings.
    \item Adjust the smile and mouth to be softer and more ambiguous, using blurred transitions at the corners to create a subtle smile, and refine the facial features, particularly the mouth and eyes, to incorporate softer transitions and avoid sharp lines, enhancing ambiguity and psychological complexity.
    \item Model the eyes with soft tonal gradations, adding subtle highlights and shadow transitions to create a direct gaze and sfumato effect.
    \item Place the hands lower in the foreground, closer to the bottom edge, forming the stable base of the figure.
    \item Render the costume with a dark dress featuring yellow detachable sleeves and a white chemise visible at the shoulder opening, avoiding excessive ornamentation.
    \item Include the rounded-backed chair with balusters, as revealed by infrared reflectography, to provide structural support for the seated pose.
    \item Add the balustrade and side columnettes to separate the foreground portrait space from the background depth, maintaining the structural integrity of the composition.
    \item Remove all jewelry, including the necklace and earrings, to maintain the absence of adornment as per the original.
\end{enumerate}
\end{tcolorbox}

\begin{tcolorbox}[
  artworkbox,
  colback=CafeBack,
  colframe=CafeFrame,
  title={\texttt{"Cafe Terrace at Night"}}
]
\begin{enumerate}[leftmargin=1.2em,itemsep=1pt,topsep=1pt]
    \item Enlarge the yellow-orange terrace and awning to dominate the left foreground and mid-ground, emphasizing their diagonal projection from the upper-left toward the center.
    \item Position a reduced-size, intensely bright yellow lantern directly beneath the awning, serving as a small but powerful light source that illuminates the terrace and cobblestones with varied warm yellow-orange reflections.
    \item Deepen the sky to a saturated ultramarine blue, adding bright white, yellow, and pale blue stars with increased density and variety, including small bright strokes in multiple sizes and colors.
    \item Move the standing waiter in white closer to the lit doorway near the center-left to serve as a vertical pivot.
    \item Enhance the cobblestone street with varied reflections of warm yellow-orange light from the terrace and lantern, deepening its color to a violet-pink and blue-violet surface using broken strokes to suggest texture.
    \item Increase the number of tables, chairs, and patrons across the terrace, clustering them densely in the lower-left and center-left under the awning.
    \item Darken the receding buildings to deep blue, violet, black-blue, and green hues, adding more orange-yellow window accents.
    \item Simplify the tree to a large, dark-green silhouette that projects into the upper-right, partly masking the right-side buildings while reducing its prominence to balance the composition.
    \item Increase the number of pedestrians in the center and right middle ground, including a variety of figures with different postures and activities.
\end{enumerate}
\end{tcolorbox}

\begin{tcolorbox}[
  artworkbox,
  colback=ScreamBack,
  colframe=ScreamFrame,
  title={\texttt{"The Scream"}}
]
\begin{enumerate}[leftmargin=1.2em,itemsep=1pt,topsep=1pt]
    \item Ensure consistent adherence to the Expressionistic style throughout the painting process, focusing on dynamic, swirling brushstrokes and strong contrasts between light and shadow.
    \item During the main-layers stage, carefully replicate the presence and positioning of all key elements such as the screaming figure, bridge railing, and background strollers while maintaining the original's emotional intensity.
    \item Avoid introducing elements not present in the original, such as detailed bridge railings or additional figures, to maintain fidelity to the source.
    \item Apply dynamic, swirling brushstrokes to the sky and other areas to capture the movement and energy of the original.
    \item Ensure that the central figure maintains the abstract, simplified form characteristic of the original, avoiding detailed anatomical features.
    \item Replicate the original's use of dynamic, horizontal lines in the bridge railing to preserve the diagonal composition and emotional tension.
    \item Apply the swirling sky technique with horizontal waves emanating from the figure, rather than circular patterns, to align with the original's emotional intensity.
    \item Use a darker, more turbulent palette for the water and distant landscape to match the original's mood of chaos and anxiety.
    \item Ensure that the central figure retains its ghostly and elongated form, emphasizing distortion and abstraction over anatomical accuracy.
    \item Replicate the original's use of swirling, dynamic brushstrokes in the sky and surrounding elements to convey movement and chaos.
    \item Keep the distant shoreline and hills minimal and abstract, maintaining the sense of isolation and focus on the central figure.
\end{enumerate}
\end{tcolorbox}

\end{tcolorbox}

\captionof{figure}{
\textbf{Artwork-specific memory $\mathcal{M}_{\mathrm{art}}$.}
The memory stores artwork-specific constraints learned during self-reflection. Each artwork name acts as a key, and the corresponding value is the set of corrective constraints used to guide later reconstruction.
}
\label{fig:mart_memory}
\vspace{-0.8em}

\clearpage
\definecolor{CodeBack}{HTML}{FFFFFF}
\definecolor{CodeFrame}{HTML}{111111}
\definecolor{UpdateGreen}{HTML}{059669}

\newcommand{\PiTransfer}{\ensuremath{\Pi^{\mathrm{*}}}}

\begin{figure}[t]
\centering
\vspace{-0.8em}

\begin{tcolorbox}[
    enhanced,
    width=\columnwidth,
    colback=CodeBack,
    colframe=CodeFrame,
    boxrule=0.55pt,
    arc=0pt,
    left=4pt,
    right=4pt,
    top=1pt,
    bottom=2pt,
    before skip=0pt,
    after skip=0pt
]

\begin{Verbatim}[
fontsize=\tiny,
commandchars=\\\{\},
breaklines=true,
breakanywhere=false,
breakautoindent=true,
breakindent=1.2em,
breaksymbolleft={},
breaksymbolright={},
baselinestretch=0.78,
tabsize=4
]
============================================================
Transfer Policy \PiTransfer
============================================================

def solver(agent, evidence_repository, artwork, policy=\PiTransfer):
    role = "art historian specializing in process reconstruction"

    schema = PEIRCEAN_SCHEMA(
        \textbf{OBSERVATION}: A directly evidenced 
        feature of the artwork or its context; 
        something that can be pointed to and verified.

        \textbf{RULE}: A general principle stating 
        that if process P were followed, observation 
        O would result.

        \textbf{HYPOTHESIS}: Therefore, process P was 
        most likely undertaken. State this as a 
        concrete, actionable instruction.

        \textbf{ACTION}: A concrete instruction 
        realizing the hypothesis; specifies what to 
        lay in or how to edit the current canvas.

        \textbf{EVIDENCE}: The evidence-repository 
        unit that grounds the observation, with 
        reliability tag: direct, indirect,
        interpretation, or speculative.
    )

    \update{method_principles = [}
        \update{"During planning, establish the overall composition and spatial relationships between elements, strictly adhering to the documented formal analysis and composition.",}
        \update{"Maintain the spatial relationships and proportions as described in the formal_analysis.composition, especially regarding the positioning of key elements like the sun and boats.",}
        \update{"Incorporate the documented materials_and_technique to guide the layering process, emphasizing the build-up of translucent washes and spontaneous brushstrokes.",}
        \update{"Adhere strictly to the monochromatic palette and avoid introducing colors or textures not present in the original.",}
        \update{"Maintain spatial compression and shallow depth throughout the composition, avoiding the use of perspective or realistic spatial cues.",}
        \update{"Focus on the dynamic interaction between elements, ensuring that each figure and object contributes to the overall narrative and emotional impact.",}
        \update{"During planning, establish the overall composition and spatial relationships between elements, strictly adhering to the documented formal analysis and composition guidelines.",}
        \update{"In the underpainting stage, prioritize capturing the subtle gradations and atmospheric effects described in the color_and_light field, rather than focusing on realistic anatomical details.",}
        \update{"Ensure that all documented elements are present and accurately represented according to their specified attributes, including shape, size, position, and texture.",}
        \update{"In the underpainting stage, prioritize capturing the subtle gradations and atmospheric effects described in the color_and_light field, rather than focusing on realism.",}
        \update{"During synthesis, ensure that the overall composition maintains the documented spatial relationships and balances, paying particular attention to the interplay between elements.",}
        \update{"In the underpainting stage, prioritize capturing the subtle gradations and atmospheric effects described in the color_and_light field, rather than focusing on realistic details.",}
        \update{"In the underpainting stage, prioritize capturing the subtle gradations and atmospheric effects described in the color_and_light field, rather than focusing on realism or additional structural elements.",}
        \update{"Maintain the monochromatic palette throughout the painting process, avoiding any unintended color shifts.",}
        \update{"During underpainting, prioritize capturing the subtle gradations and atmospheric effects described in the color_and_light field, rather than focusing on realism.",}
        \update{"In the planning stage, establish the overall composition and spatial relationships between elements, strictly adhering to the documented formal analysis and composition.",}
        \update{"Ensure the application of impasto techniques throughout the painting process to mimic the textured surface of the original.",}
        \update{"Set the palette strictly according to the color_and_light field, prioritizing bold, contrasting hues over softer, more muted tones.",}
        \update{"Maintain the structural composition while adhering closely to the formal_analysis.composition and proportion fields to preserve the intended balance and focus.",}
        \update{"During the main-layers stage, apply multiple translucent glazes to build up color depth and texture, mimicking the original's material richness."}
    \update{]}

    constraints = (
        "1. Apply Peircean schema for every step: "
        "observation, rule, hypothesis, action.\textbackslash n"
        "2. Every step must cite at least one specific key "
        "from the evidence JSON.\textbackslash n"
        "3. Use method_principles to guide reconstruction "
        "when the target artwork has sparse evidence.\textbackslash n"
        "4. Steps must appear in production order; "
        "step 1 is first, step N is last.\textbackslash n"
        "5. Return valid JSON only.\textbackslash n"
    )
    
    return trajectory
\end{Verbatim}

\end{tcolorbox}

\caption{
\textbf{Transferred policy $\Pi^{\mathrm{*}}$ for non-canonical artworks.}
The policy keeps the base Peircean abductive structure while incorporating learned method-principles from canonical artworks. These principles guide reconstruction when the target artwork has limited local evidence.
}
\label{fig:transfer_policy}
\vspace{-0.8em}

\end{figure}
\begin{figure*}[t]
\centering
\includegraphics[width=\textwidth]{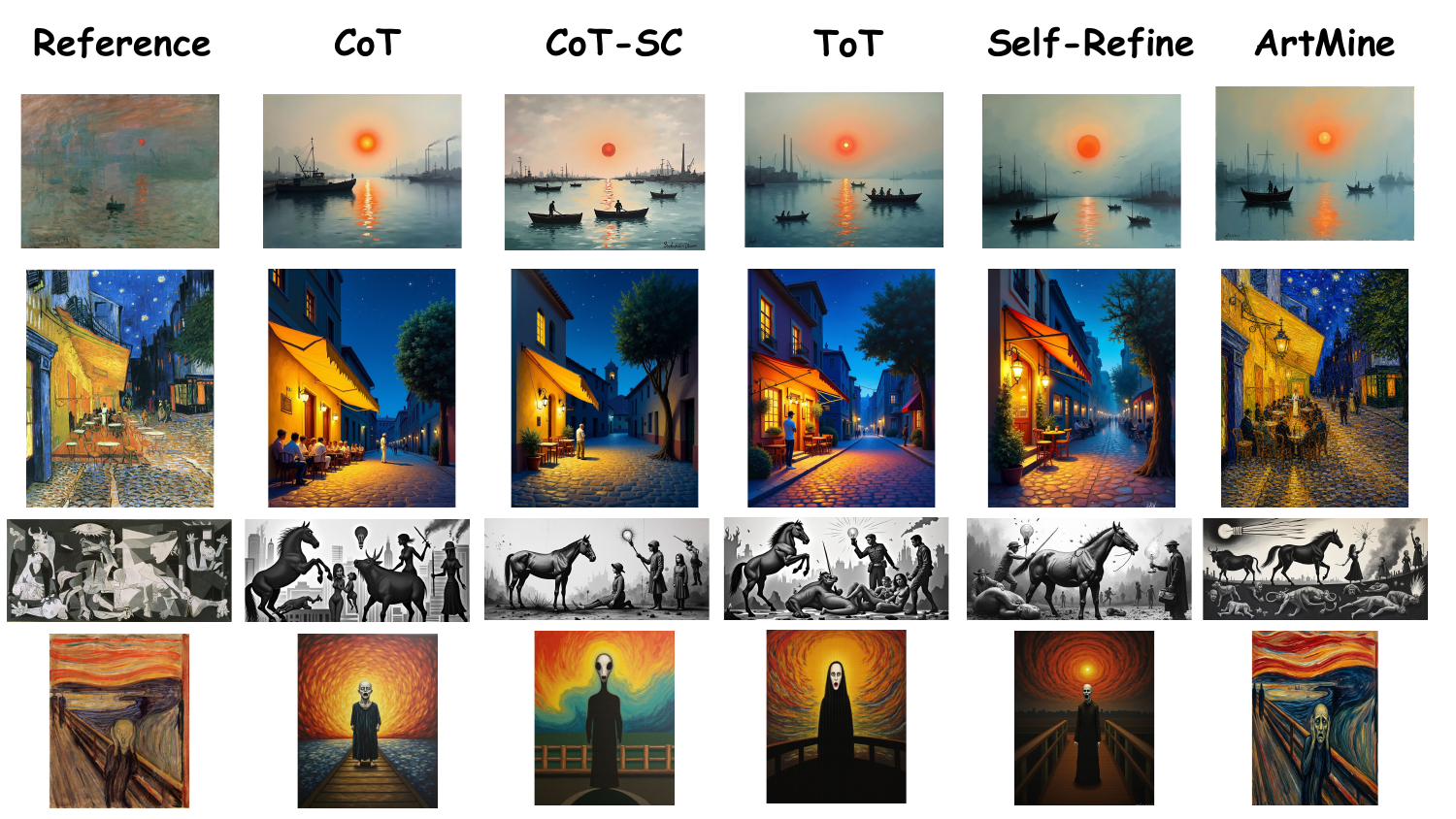}
\caption{
\textbf{Qualitative comparison of generated reconstructions across canonical artworks.}
Each row corresponds to one canonical artwork: \textit{Impression, Sunrise} by Claude Monet, \textit{Caf\'e Terrace at Night} by Vincent van Gogh, \textit{Guernica} by Pablo Picasso, and \textit{The Scream} by Edvard Munch. For each artwork, we show the reference image followed by reconstructions from CoT, CoT-SC, ToT, Self-Refine, and ArtMine. Compared with baseline prompting strategies, ArtMine more consistently preserves global composition, foreground structure, motif placement, and painterly treatment.
}
\label{fig:qualitative_comparison_seven_artworks}
\end{figure*}
\begin{figure*}[t]
\centering
\includegraphics[width=\textwidth]{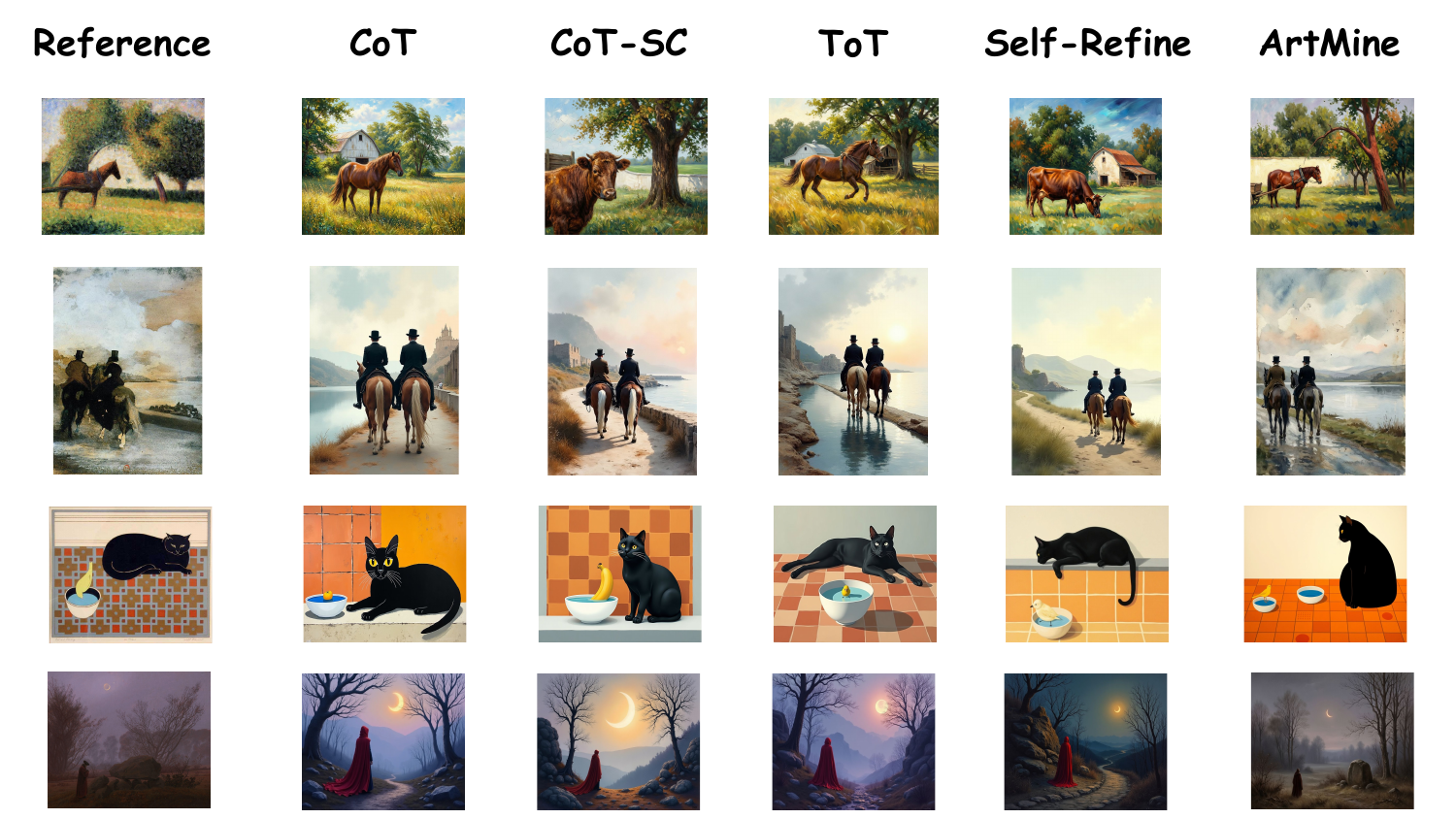}
\caption{
\textbf{Qualitative comparison of generated reconstructions across non-canonical artworks.}
Each row corresponds to one non-canonical artwork: \textit{Horse and Cart} by Georges Seurat, \textit{Two Riders by a River} by Edgar Degas, \textit{Cat and Canary} by Will Barnet, and \textit{A Walk at Dusk} by Caspar David Friedrich. For each artwork, we show the reference image followed by reconstructions from CoT, CoT-SC, ToT, Self-Refine, and ArtMine. Compared with baseline prompting strategies, ArtMine more consistently preserves global composition, foreground structure, motif placement, and painterly treatment.
}
\label{fig:qualitative_comparison_non_canon_artworks}
\end{figure*}
\begin{figure*}[t]
    \centering

    \includegraphics[
        width=\textwidth,
        keepaspectratio
    ]{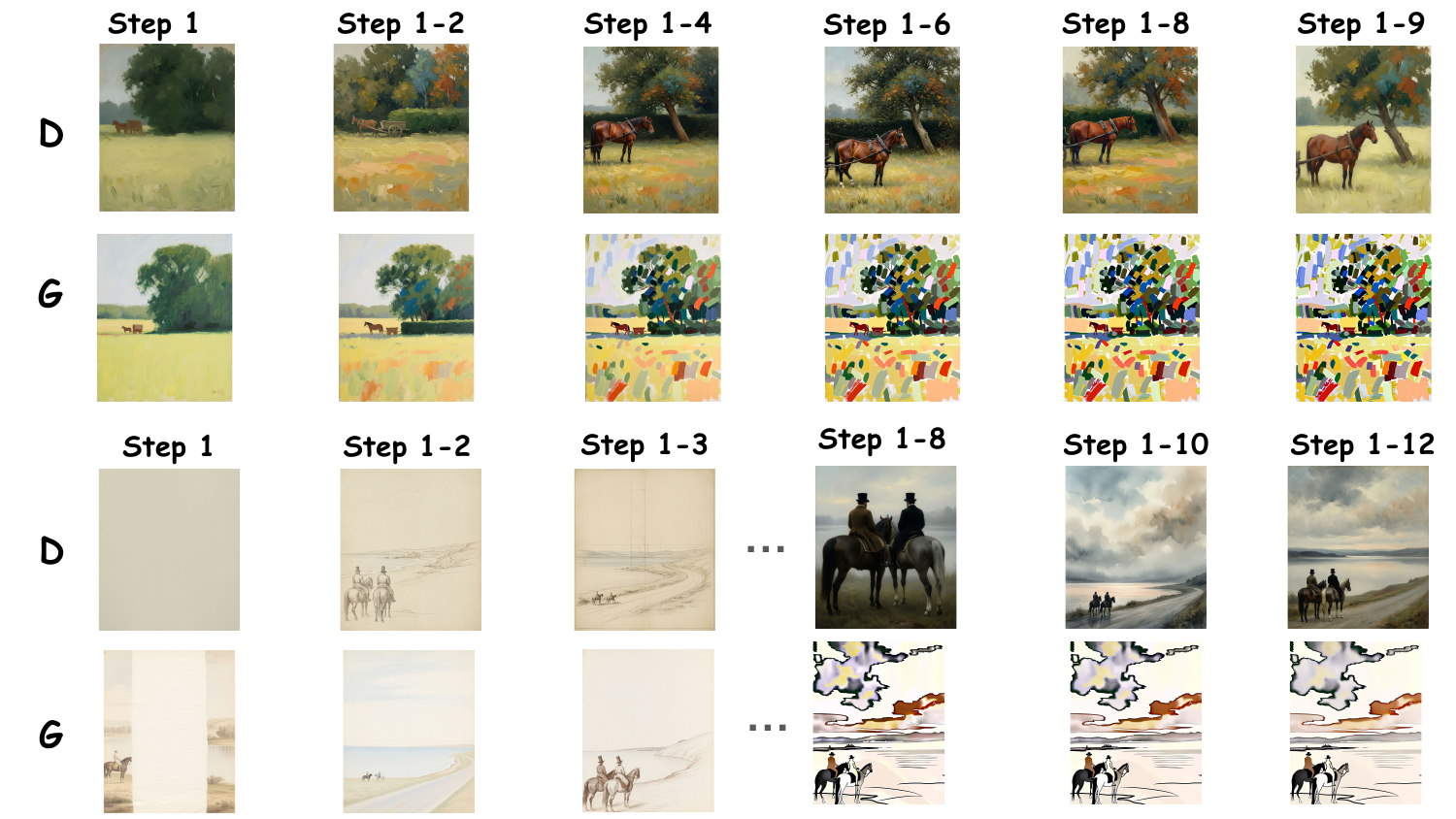}

    \vspace{0.5em}

    \includegraphics[
        width=\textwidth,
        keepaspectratio
    ]{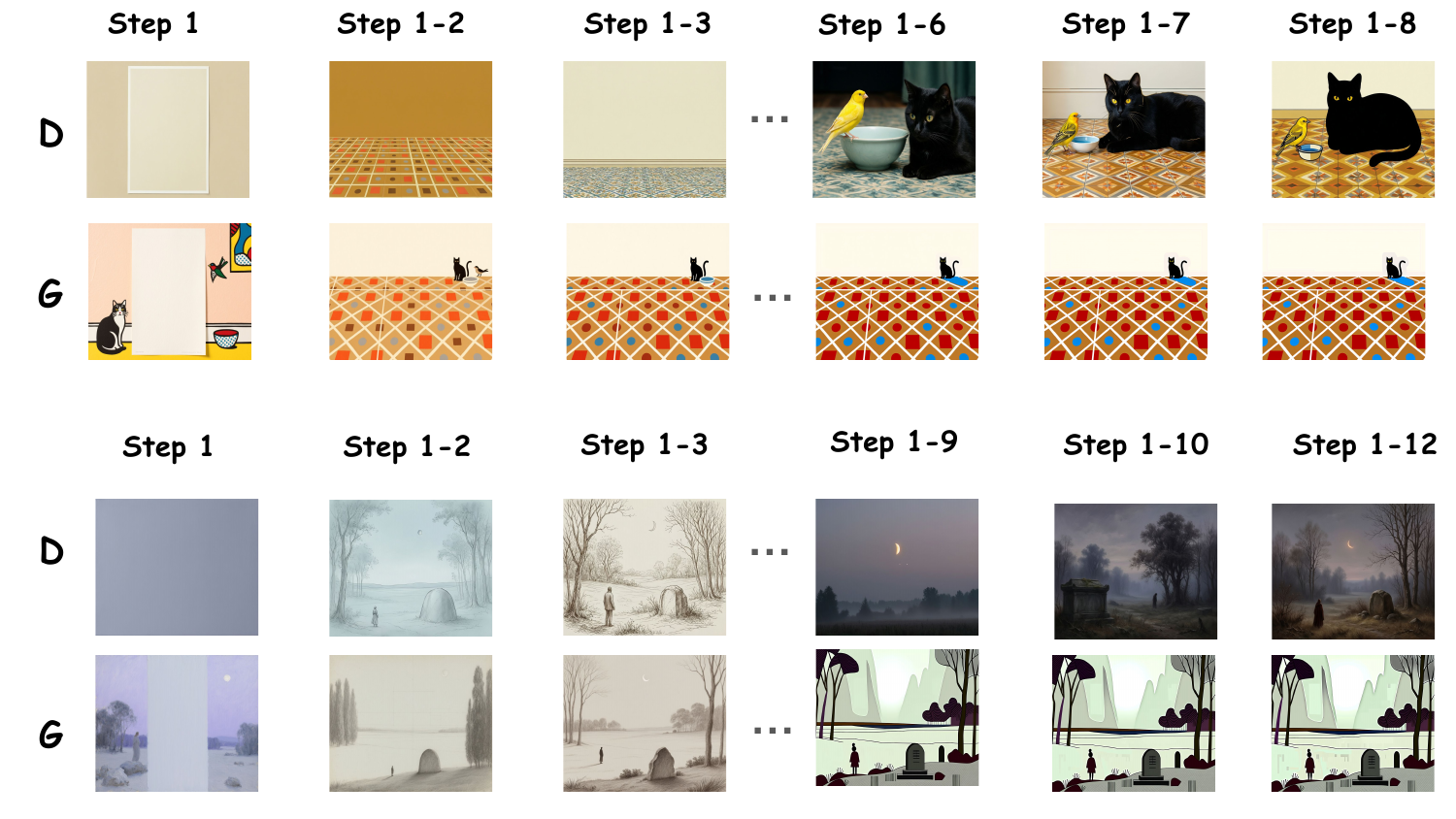}

  \caption{
\textbf{Direct and guided staged reconstructions across non-canonical artworks.}
For each artwork, \textbf{D} denotes direct stage-wise generation, where each stage is rendered independently from its partial process representation, while \textbf{G} denotes guided stage-wise generation, where each stage is conditioned on the previously generated image. Long trajectories are compactly shown using the first three and last three stages. Across the non-canonical artworks---\textit{Horse and Cart} by Georges Seurat, \textit{Two Riders by a River} by Edgar Degas, \textit{Cat and Canary} by Will Barnet, and \textit{A Walk at Dusk} by Caspar David Friedrich; we observe complementary failure modes. Direct generation completes the artwork too early because each stage is rendered as a near-final image, whereas guided generation better preserves stage-to-stage continuity but can accumulate errors and drift in style over later stages.
}
    \label{fig:noncanonical_direct_guided_stages}
\end{figure*}

\end{document}